\documentclass[journal]{IEEEtran}
\usepackage{graphicx}
\usepackage{cite}
\usepackage{epstopdf}
\usepackage{amsfonts}
\usepackage{amsmath}
\usepackage{algorithm}
\usepackage{algorithmic}
\usepackage{subfig}
\usepackage{array}
\usepackage{multicol}
\usepackage{multirow}
\usepackage{enumerate}
\begin{document}

\title{Deep Image-to-Video Adaptation and Fusion Networks for Action Recognition}

\author{Yang~Liu,  Zhaoyang~Lu,~\IEEEmembership{Senior~Member,~IEEE,} Jing Li,~\IEEEmembership{Member,~IEEE,} Tao Yang,~\IEEEmembership{Member,~IEEE,}  and~Chao~Yao
\thanks{This work is supported in part by the National Natural Science Foundation of China (Nos. 61802315, 61502364, 61672429), and Natural Science Foundation of Shaanxi Province (No. 2018JQ6013). (\emph{Corresponding author: Jing Li and Chao Yao.})}
\thanks{Yang Liu is with the School of Data and Computer Science, Sun Yat-Sen University, Guangzhou 510006, China, and also with the School of Telecommunications Engineering, Xidian University, Xi'an 710071, China (e-mail: liuy856@mail.sysu.edu.cn).}
\thanks{Zhaoyang Lu and Jing Li are with the School of Telecommunications Engineering, Xidian University, Xi'an 710071, China (e-mail: zhylu@xidian.edu.cn; jinglixd@mail.xidian.edu.cn).}
\thanks{Tao Yang is with the School of Computer Science, Northwestern Polytechnical University, Xi'an 710072, China (e-mail: tyang@nwpu.edu.cn).}
\thanks{Chao Yao is with the School of Automation, Northwestern Polytechnical University, Xi'an 710072, China (e-mail: yaochao@nwpu.edu.cn).}}

\markboth{MANUSCRIPT SUBMITTED TO IEEE TRANSACTIONS ON IMAGE PROCESSING}%
{Shell \MakeLowercase{\textit{et al.}}: Bare Demo of IEEEtran.cls for IEEE Journals}

\maketitle

\begin{abstract}
Existing deep learning methods for action recognition in videos require a large number of labeled videos for training, which is labor-intensive and time-consuming. For the same action, the knowledge learned from different media types, e.g., videos and images, may be related and complementary. However, due to the domain shifts and heterogeneous feature representations between videos and images, the performance of classifiers trained on images may be dramatically degraded when directly deployed to videos without effective domain adaptation and feature fusion methods. In this paper, we propose a novel method, named Deep Image-to-Video Adaptation and Fusion Networks (DIVAFN), to enhance action recognition in videos by transferring knowledge from images using video keyframes as a bridge. The DIVAFN is a unified deep learning model, which integrates domain-invariant representations learning and cross-modal feature fusion into a unified optimization framework. Specifically, we design an efficient cross-modal similarities metric to reduce the modality shift among images, keyframes and videos. Then, we adopt an autoencoder architecture, whose hidden layer is constrained to be the semantic representations of the action class names. In this way, when the autoencoder is adopted to project the learned features from different domains to the same space, more compact, informative and discriminative representations can be obtained. Finally, the concatenation of the learned semantic feature representations from these three autoencoders are used to train the classifier for action recognition in videos. Comprehensive experiments on four real-world datasets show that our method outperforms some state-of-the-art domain adaptation and action recognition methods.
\end{abstract}

\begin{IEEEkeywords}
Action recognition, adaptation, deep learning, fusion.
\end{IEEEkeywords}

\IEEEpeerreviewmaketitle

\section{Introduction}
\IEEEPARstart{R}{ecent} works \cite{DBLP:conf/cvpr/KarpathyTSLSF14,DBLP:conf/nips/SimonyanZ14,DBLP:conf/iccv/TranBFTP15,DBLP:journals/spl/LiuLLYY18} show that deep convolutional neural networks (CNNs) are promising for action recognition in videos. Since a prerequisite for a well-trained deep model is the availability of large amounts of labeled training videos, the collection, preprocessing,  and annotation of such datasets (e.g. HMDB51 \cite{DBLP:conf/iccv/KuehneJGPS11}, UCF101 \cite{DBLP:journals/corr/abs-1212-0402}, ActivityNet \cite{DBLP:conf/cvpr/HeilbronEGN15}, Something Something \cite{DBLP:conf/iccv/GoyalKMMWKHFYMH17}, AVA \cite{DBLP:conf/cvpr/GuSRVPLVTRSSM18}, Moments in Time \cite{8651343}) is often labor-intensive and time-consuming. Moreover, storing and training on such large amounts of video data can consume substantial computational resources \cite{DBLP:journals/pr/MaBZSS17}. In some extreme conditions, it is often difficult and infeasible to capture adequate number of videos. Unlike videos, images are much easier and cheaper to be collected and annotated, and there also exists many labeled image datasets which can be utilized such as the BU101 \cite{DBLP:journals/pr/MaBZSS17}, Stanford40 \cite{DBLP:conf/iccv/YaoJKLGF11}, DB \cite{DBLP:journals/pami/GuptaKD09}, PPMI \cite{DBLP:conf/cvpr/YaoF10}, Willow-Actions \cite{DBLP:conf/bmvc/DelaitreLS10}, Still DB \cite{DBLP:conf/icpr/IkizlerCPD08}, HII \cite{DBLP:conf/mm/LiWZK17} and VOC 2012 \cite{Everingham2010}, etc. Moreover, the computational cost of learning deep models of images is much less than that of videos \cite{DBLP:conf/ijcai/YuWSD18}. Therefore, the video data are rarer than the image data due to the high acquisition cost and the hard annotating \cite{DBLP:conf/ijcai/YuWSD18,DBLP:journals/tcyb/ZhangHTHJ17,8578655}.

\begin{figure}[!t]
\centering
\includegraphics[scale=0.72]{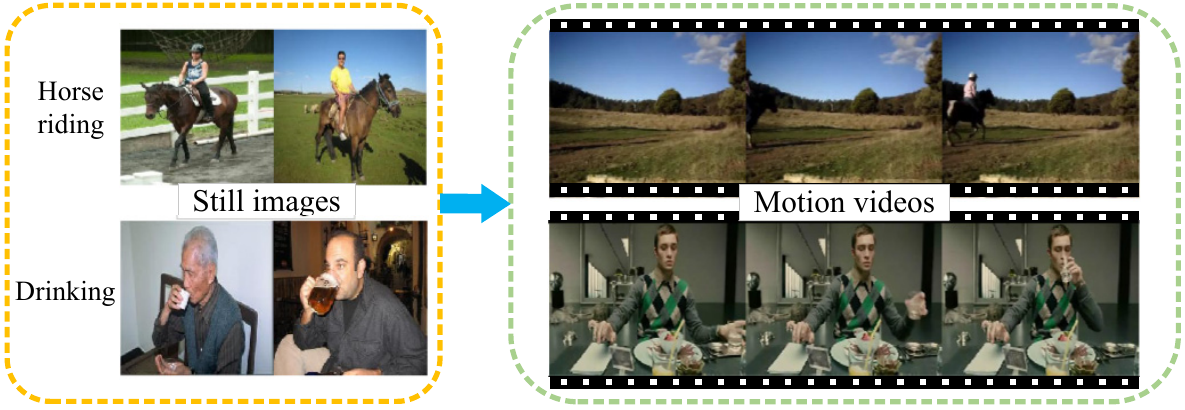}
\caption{Illustration of still images and motion videos of categories horse riding and drinking.  }
\label{Fig1}
\end{figure}

In addition, images tend to focus on representative moments and have more diverse examples of an action in terms of camera viewpoint, background, human appearance, posture, etc. In contrast, videos usually capture a whole temporal progression of an action and contain many redundant and uninformative frames (e.g. action classes horse riding and biking in still images and motion videos, shown in Figure \ref{Fig1}). For example, one may need to look through all, or most, of the video frames to determine an action class in the video, while a single glance is enough to annotate the action in the image. This fact makes image data a good auxiliary source to enhance video action recognition. Furthermore, semantically related images and videos may have the similar appearance, posture and object, as shown in Figure \ref{Fig1}. If the inherent semantic relationship between images and videos can be used to fuse the features from image and video data, the performance of the action recognition in videos could be improved and we can use fewer labeled video action data and achieve comparable performance to the traditional ones which use much more labeled training video samples.

However, directly discriminating the video data by the classifier trained on images will result in degraded performance \cite{DBLP:conf/eccv/SaenkoKFD10}. As shown in Figure \ref{Fig1}, the still images (left) are different from video frames (right) in viewpoint, background, appearance, posture, etc. In addition, video features are often represented in spatiotemporal form, which is significantly different from image representations in terms of feature dimensionality and physical meaning. Nevertheless, the action information between image and video is complementary, which was validated by works \cite{DBLP:journals/pr/MaBZSS17,8060555,8578655}. And previous cross-domain action recognition related works \cite{7432005,7364249,DBLP:journals/complexity/0084LLYD18,8453034} validated that utilize the action knowledges from other domains can substantially improve the performance of action recognition in the target domain. Based on these observations, in this paper, we focus on enhancing action recognition in videos with limited training samples by utilizing the image data, meanwhile solving the aforementioned domain shift problem. In general, we propose a novel unified deep learning framework, named Deep Image-to-Video Adaption and Fusion Networks (DIVAFN), which learns domain-invariant representations and fuses cross-modal feature, simultaneously. An overview of the DIVAFN is presented in Figure \ref{Fig2}. Since a video consists of a sequence of frames which are essentially a collection of images, a video is correlated with its related keyframes \cite{DBLP:journals/tcsv/LuoPC09,8578867}. Therefore, we can explore the relationship between images and videos by using video keyframes as the bridge.

The DIVAFN consists of three deep neural networks, one for image modality, one for video keyframe modality, and the other for video modality. To reduce the domain shift among images, keyframes and videos, we design a novel cross-modal similarities metric. Because of the similarity on the semantic representation of the same action from the data in different modalities, we utilize the semantic relationship among images, keyframes, and videos as a guidance to fuse the features learned from these domains. Inspired by the semantic autoencoder \cite{DBLP:conf/cvpr/KodirovXG17} proposed for zero-shot learning, we adopt this autoencoder architecture with a constraint that the latent representations from the hidden layer of the autoencoder should be equal to the semantic representations of the action class names (e.g. attribute representations \cite{DBLP:conf/cvpr/LampertNH09} or word2vec representations \cite{DBLP:conf/nips/MikolovSCCD13}). With this constraint, the autoencoder can project the learned domain-invariant features to the semantic space, and the projected semantic representations can also be projected back to the domain-invariant feature space. Since the images, keyframes and videos from the same action class have the same semantic meaning, we can obtain more compact, informative and discriminative feature representations from the hidden layer of the autoencoder, which contains both learned domain-invariant feature knowledge and semantic relationship knowledge. Different from previous work \cite{DBLP:conf/cvpr/KodirovXG17} using semantic autoencoders to solve zero-shot learning problems, we unify the semantic autoencoder and domain-invariant learning into a unified deep learning framework for image-to-video adaption problem. In more detail, we simultaneously project the learned domain-invariant features from keyframes, videos and their concatenations to the same semantic space by learning three semantic autoencoders. Then, the concatenation of the learned semantic feature representations from these three autoencoders is used to train the classifier for action recognition in videos. Extensive experiments on four real-world datasets show that our approach significantly outperforms state-of-the-art approaches.

The main contributions of this paper are as follows:
\begin{itemize}
\item To perform domain-invariant representations learning, we propose a novel unified learning framework with three deep neural networks, one for image modality, one for keyframe modality, and the other for video modality, and design a novel cross-modal similarities metric to reduce the modality shift among images, keyframes and videos.
\item Since the same action class for both image and video action data share the same semantic information, we utilize the semantic information as a guide to fuse the images, keyframes and videos features after learning domain-invariant features. An autoencoder architecture with a constraint that the representations from hidden layer should be equal to the semantic representations of the action classes is adopted. In this way, we can obtain more compact, informative and discriminative representations, which contains both learned domain-invariant feature knowledge and semantic relationship knowledge.
\item To effectively fuse keyframe and video feature representations, we simultaneously project the learned domain-invariant keyframe features, video features and their concatenations to the same semantic space by learning three semantic autoencoders. And the concatenation of the learned semantic representations from these three autoencoders are used to train the classifier for action recognition in videos. Extensive experimental results validate the effectiveness of the newly designed representations.
\item One major advantage of the proposed method is that it takes the situation when the number of training videos is limited into consideration. We transfer knowledge from images to videos, and integrate domain-invariant representations learning and feature fusion into a unified deep learning framework, named Deep Image-to-Video Adaptation and Fusion Networks (DIVAFN).
\end{itemize}

This paper is organized as follows: Section \uppercase\expandafter{\romannumeral2} briefly reviews the related works. Section \uppercase\expandafter{\romannumeral3} introduces the proposed DIVAFN. Experimental results and related discussions are presented in Section \uppercase\expandafter{\romannumeral4}. Finally, Section \uppercase\expandafter{\romannumeral5} concludes the paper.

\section{Related Work}
\subsection{Action Recognition on Videos}
Action recognition is a very active research field and has received great attention in recent years \cite{DBLP:journals/csur/AggarwalR11}. Conventional action recognition methods mainly consist of two steps: feature extraction and classifier training. Many hand-crafted features are designed to capture spatiotemporal information in videos. Some representative works include space-time interest points \cite{DBLP:conf/cvpr/LaptevMSR08}, dense trajectories \cite{DBLP:journals/ijcv/WangKSL13} and Improved Dense Trajectories (IDT) \cite{DBLP:conf/iccv/WangS13a}. Advance feature encoding methods such as Fisher vector (FV) encoding \cite{DBLP:conf/eccv/PerronninSM10}, Vector of Locally Aggregated Descriptors (VLAD) encoding \cite{DBLP:conf/cvpr/JegouDSP10} and Locality-constrained Linear Coding (LLC) encoding \cite{DBLP:conf/cvpr/WangYYLHG10}, can further improve the performance of these features. Some recent works use deep neural networks (particularly CNNs) to jointly learn the feature extractors and classifiers for action recognition in videos. Simonyan and Zisserman \cite{DBLP:conf/nips/SimonyanZ14} proposed two-stream CNNs: one stream captures the spatial information from video frames and the other steam captures motion information from stacked optical flows. Tran et al. \cite{DBLP:conf/iccv/TranBFTP15} and Ji et al. \cite{DBLP:journals/pami/JiXYY13} proposed 3D convolutional networks to learn spatiotemporal features. Wang et al. \cite{DBLP:conf/cvpr/Wang0T15} extracted trajectory-pooled deep-convolutional descriptor (TDD) using deep architectures to learn discriminative convolutional feature maps, which are then aggregated into effective descriptors by trajectory-constrained pooling. However, these deep learning methods all require a large amount of annotated videos to avoid overfitting.

To address the limited training videos problem, our approach enhances action recognition in videos by transferring knowledge from images, and integrate domain-invariant representations learning and cross-modal feature
fusion into a unified deep learning framework.

\subsection{Image-to-Video Action Recognition}
To harness the knowledge from images to enhance action recognition in videos, several recent works have focused on knowledge adaptation from images to videos and achieve good performance in image-to-video action recognition \cite{DBLP:conf/eccv/GanSDG16,DBLP:journals/tcyb/ZhangHTHJ17,DBLP:conf/mm/LiWZK17,DBLP:journals/pr/MaBZSS17,DBLP:conf/ijcai/YuWSD18,DBLP:journals/tip/YuWCD19}. Ma et al. \cite{DBLP:journals/pr/MaBZSS17} used both images and videos to jointly train shared CNNs and use the shared CNNs to map images and videos into the same feature space. They verified that images are complementary to videos by extensive experimental evaluation. Yu et al. \cite{DBLP:conf/ijcai/YuWSD18} proposed a Hierarchical Generative Adversarial Networks (HiGANs) based image-to-video adaptation method by leveraging the relationship among images, videos and their corresponding video frames. Gan et al. \cite{DBLP:conf/eccv/GanSDG16} proposed a mutually voting strategy to filter noisy images and video frames and then jointly explored images and videos to address labeling-free video recognition. Zhang et al. \cite{DBLP:journals/tcyb/ZhangHTHJ17} proposed a classifier based image-to-video adaptation method by exploring and utilizing the common knowledge of both labeled videos and images. Li et al. \cite{DBLP:conf/mm/LiWZK17} used class-discriminative spatial attention maps to transfer knowledge from images to videos. Yu et al. \cite{DBLP:journals/tip/YuWCD19} proposed a symmetric generative adversarial learning approach (Sym-GANs) to learn domain-invariant augmented feature with excellent transferability and distinguishability for heterogeneous image-to-video adaptation. However, these methods usually require sufficient labeled training data for both images and videos to learn domain-invariant features. In these methods, the semantic relationship between images and videos is also ignored. In addition, the evaluation datasets used in these works is relatively simple and small compared to the datasets used in our work. For instance, BU101$\rightarrow$UCF101 are the image and video datasets used in our study. The BU101, which contains $23.8$K images of $101$ action classes, is the first action image dataset that has one-to-one correspondence in action classes with UCF101, which is a large-scale action recognition video benchmark dataset. To meet the challenge of the image-video datasets in our study, more specific and effective domain adaption and cross-modal feature fusion methods should be designed to reduce the significantly larger modality shift problem, and realize the fusion of the complementary information between images and video, respectively.

In this paper, by taking the semantic information between images and videos into consideration, we develop a new method to recognize the actions in video when the number of training videos is limited. To utilize the semantic relationship between images and videos, we adopt an autoencoder architecture with a constraint that the representations from hidden layer should be equal to the semantic representations of the action class names. In this way, the semantic information is employed, which makes the number of labeled training samples reduced in our method.
\subsection{Domain Adaptation on Heterogeneous Features}
Domain shift refers to the situation where heterogeneous feature representations exists between source and target domains, which may cause the classifier learned from source domain to perform poorly on target domain. A large number of domain adaptation approaches have been proposed to address this problem \cite{DBLP:conf/iccv/LongWDSY13,DBLP:conf/cvpr/ZhangLO17,DBLP:conf/icml/LongC0J15,8502831,DBLP:conf/cvpr/JiangL17}. Long et al. \cite{DBLP:conf/iccv/LongWDSY13} proposed a distribution adaptation method to find a common subspace where the marginal and conditional distribution shifts between domains are reduced. Zhang et al. \cite{DBLP:conf/cvpr/ZhangLO17} learned two projections to map the source and target domains into their respective subspaces where both the geometrical and statistical distribution difference are minimized. Long et al. \cite{DBLP:conf/icml/LongC0J15} built a deep adaptation network (DAN) that explores multiple kernel variant of maximum mean discrepancies (MK-MMD) to learn transferable features. Zhang et al. \cite{8502831} explored the underlying complementary information from multiple views and proposed a novel subspace clustering model for multi-view data named Latent Multi-View Subspace Clustering (LMSC). Jiang et al. \cite{DBLP:conf/cvpr/JiangL17} proposed a deep cross-modal hashing method to map images and texts into a domain-invariant hash space, and then performed similarity search in multimedia retrieval applications. Huang et al. \cite{DBLP:conf/cvpr/HuangP18} jointly minimized the media-level and correlation-level domain discrepancies across texts and images to enrich the training information and boost the retrieval accuracy on target domain. In most of these works \cite{DBLP:conf/iccv/LongWDSY13,DBLP:conf/cvpr/ZhangLO17,DBLP:conf/icml/LongC0J15,8502831}, data from source and target domains belonged to the same media type (videos, images or texts), and other works \cite{DBLP:conf/cvpr/JiangL17}, \cite{DBLP:conf/cvpr/HuangP18} focused on image-text retrieval problem.

In contrast, our method focuses on heterogeneous image-to-video action recognition problem where the video is represented by spatial-temporal feature that totally differs from the image representation in terms of feature dimensionality and physical meaning. To perform domain-invariant representations learning, we propose a unified learning framework with three deep neural networks, which are used for images, videos, and keyframes, individually, and design a new cross-modal similarities metric to reduce the modality shift among them.

\section{Deep Image-to-Video Adaptation and Fusion}

\begin{figure*}[!ht]
\centering
\includegraphics[scale=0.21]{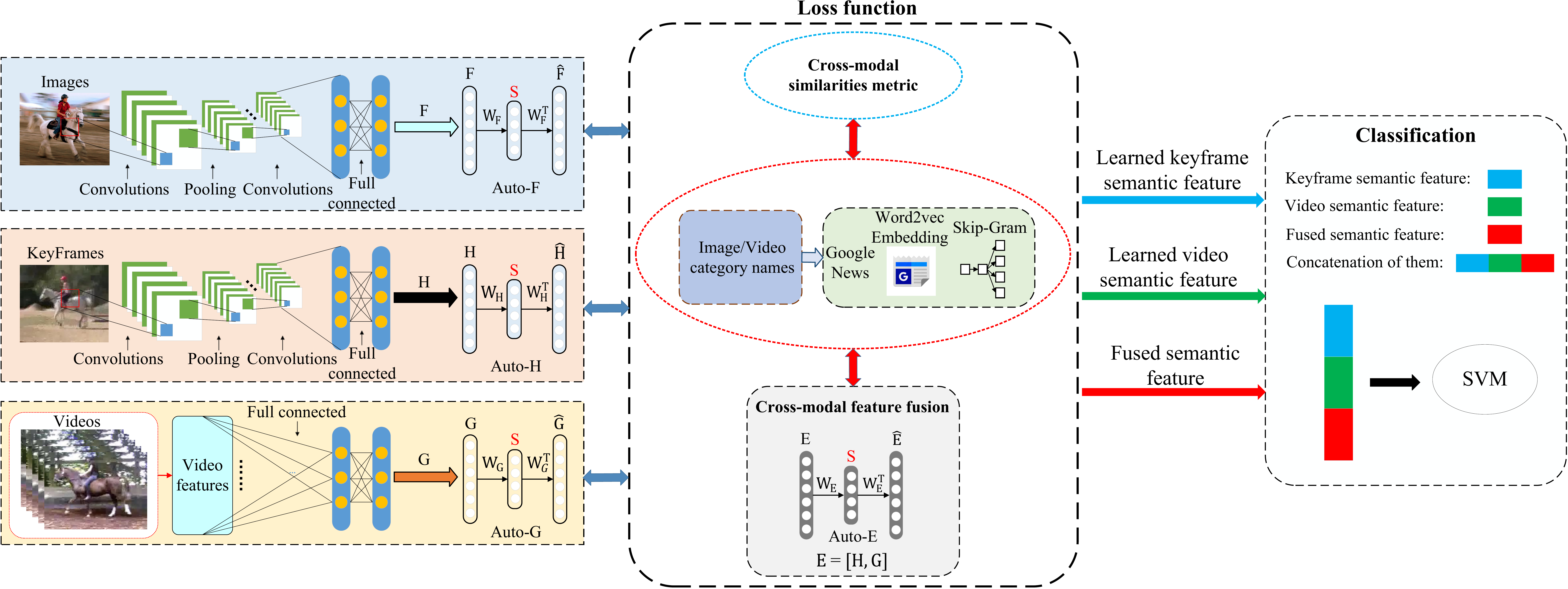}
\caption{Framework of our proposed method DIVAFN.  }
\label{Fig2}
\end{figure*}

\subsection{Framework Overview}
The framework of the DIVAFN is shown in Figure \ref{Fig2}, which is a unified deep learning framework seamlessly constituted by two parts: the domain-invariant representations learning one and the cross-modal feature fusion one. The domain-invariant representations learning part contains three deep neural networks, one for image modality, one for keyframe modality, and the other for video modality. Considering the computational efficiency, we extract keyframes from videos by histogram difference based method \cite{DBLP:journals/tcyb/ZhangHTHJ17}. The inputs of the image and keyframe networks are raw image pixels, while the inputs of the video network is video features such as the Improved Dense Trajectories (IDT) \cite{DBLP:conf/iccv/WangS13a} and the C3D features \cite{DBLP:conf/iccv/TranBFTP15}. To learn domain-invariant representations, we design a novel cross-modal similarities metric to reduce the domain shift among images, keyframes and videos. With these domain-invariant features, we utilize the semantic relationship among images, keyframes and videos to perform cross-modal feature fusion. Specifically, we adopt an autoencoder architecture and exert an additional constraint that the latent representations from the hidden layer of the autoencoder should be equal to the semantic representations of the action class names (e.g. attribute representations \cite{DBLP:conf/cvpr/LampertNH09} or word2vec representations \cite{DBLP:conf/nips/MikolovSCCD13}). Then, we simultaneously project the learned domain-invariant keyframe features, video features and their concatenations to the same semantic space by learning three semantic autoencoders. Finally, the concatenation of the learned semantic feature representations from these three autoencoders is used to train the Support Vector Machine (SVM) \cite{DBLP:journals/tist/ChangL11} classifier for action recognition in videos.
\subsection{Notations}
To fix notation, boldface lowercase letters like $\mathbf{w}$ denote vectors. Boldface uppercase letters like $\mathbf{W}$ denote matrices, and the element in the $i$th row and $j$th column of matrix $\mathbf{W}$ is denoted as $W_{ij}$. The $i$th row of $\mathbf{W}$ is denoted as $\mathbf{W}_{i*}$, and the $j$th column of $\mathbf{W}$ is denoted as $\mathbf{W}_{*j}$. Assume that we have $n$ training samples and each training sample has image, keyframe and video modalities. We use $\mathbf{X}=\{\mathbf{x}_i\}_{i=1}^n$ to denote the image modality, where $\mathbf{x}_i$ can be handcrafted features or the raw pixels of image $i$. And we use $\mathbf{Y}=\{\mathbf{y}_j\}_{j=1}^n$ to denote the keyframe modality, where $\mathbf{y}_j$ can be handcrafted features or the raw pixels of keyframe $j$. Moreover, we use $\mathbf{Z}=\{\mathbf{z}_k\}_{k=1}^n\in \mathbb{R}^{d_v\times n}$ to denote the video modality, where $\mathbf{z}_k$ is visual feature for video $k$, $d_v$ is the dimension of the input video feature. In addition, we define three cross-modal similarity matrices $\textbf{M}^1$, $\textbf{M}^2$ and $\textbf{M}^3$. If image $\mathbf{x}_i$ and video $\mathbf{z}_k$ belong to the same action class, $M^1_{ik}=1$. Otherwise, $M^1_{ik}=0$. If image $\mathbf{x}_i$ and keyframe $\mathbf{y}_j$ belong to the same action class, $M^2_{ij}=1$. Otherwise, $M^2_{ij}=0$. If keyframe $\mathbf{y}_j$ and video $\mathbf{z}_k$ belong to the same action class, $M^3_{jk}=1$. Otherwise, $M^3_{jk}=0$. Let $f(\mathbf{x}_i; \theta_x) \in \mathbb{R}^d$, $h(\mathbf{y}_j; \theta_y) \in \mathbb{R}^d$,  $g(\mathbf{z}_k; \theta_z) \in \mathbb{R}^d$ denote the learned domain-invariant feature for the $i$-th image, $j$-th keframe, and $k$-th video, individually, where $d$ denotes the dimension of the features. The parameters of the CNNs for image, keyframe and video modalities are defined as $\theta_x$,   $\theta_y$ and $\theta_z$, respectively. The semantic representations of action class names for image, keyframe and video modalities are denoted as $\mathbf{S}_i\in \mathbb{R}^{k\times n}$,  $\mathbf{S}_f\in \mathbb{R}^{k\times n}$ and $\mathbf{S}_v\in \mathbb{R}^{k\times n}$, respectively, where $k$ is the dimension of semantic representation.

\subsection{Model Formulation}
\subsubsection{Network Architecture}
The deep neural network for both image modality and keyframe modality is a convolutional neural network CNN-F adapted from \cite{DBLP:conf/bmvc/ChatfieldSVZ14}. The CNN-F has been trained on the ImageNet $2012$ and it consists of five convolutional layers (conv1-conv5) and three fully connected layers (fc6, fc7, fc8).  The input of our network is the raw image pixels. The first seven layers of our network are the same as those in CNN-F \cite{DBLP:conf/bmvc/ChatfieldSVZ14}. The eighth layer is a fully connected layer with the output being the learned domain-invariant features. All the first seven layers use the Rectified Linear Unit (ReLU) \cite{DBLP:conf/nips/KrizhevskySH12} as activation function. For the eighth layer, we choose identity function as the activation function.

To perform domain-invariant feature learning for video modality, we first represent each video $\mathbf{y}_j$ as a vector with Locality-constrained Linear Coding (LLC) \cite{DBLP:conf/cvpr/WangYYLHG10} encoding of Improved Dense Trajectories (IDT) \cite{DBLP:conf/iccv/WangS13a}, or the Convolutional 3D (C3D) visual features provided by \cite{DBLP:conf/iccv/TranBFTP15}. And then the extracted video vectors are used as the input to a deep neural network with two fully connected layers. The activation function for the first layer is ReLU and that for the second layer is the identity function. The detailed configuration of the deep neural network for image, keyframe and video modalities can be seen on the website https://yangliu9208.github.io/DIVAFN/. We choose these deep networks because they have been validated their effectiveness in the previous work \cite{DBLP:conf/cvpr/VenkateswaraECP17}.

 \subsubsection{Domain-invariant Representations Learning and Fusion}
We first demonstrate how to exploit the label knowledge from training images, video keyframes and videos by designing a cross-modal similarities metric. Since the inner product $\langle\cdot,\cdot\rangle$ has been validated as a good pairwise similarity measure by previous work \cite{DBLP:conf/cvpr/JiangL17,DBLP:conf/iccv/CaoLWY17}, we use the inner product $\langle\mathbf{F},\mathbf{G}\rangle$,  $\langle\mathbf{F},\mathbf{H}\rangle$ and $\langle\mathbf{H},\mathbf{G}\rangle$ as similarity measures for domain-invariant representations learning, where $\mathbf{F}\in\mathbb{R}^{d\times n}$ with $\mathbf{F}_{*i}=f(\mathbf{x}_i;\theta_x)$ is the output matrix of the image deep neural network, $\mathbf{H}\in\mathbb{R}^{d\times n}$ with $\mathbf{H}_{*j}=h(\mathbf{y}_j;\theta_y)$ is the output matrix of the keyframe deep neural network, $\mathbf{G}\in\mathbb{R}^{d\times n}$ with $\mathbf{G}_{*k}=g(\mathbf{z}_k;\theta_z)$ is the output matrix of the video deep neural network, $d$ is the dimension of the learn domain-invariant features, and $n$ is the number of training samples. Larger value of the inner product means high similarity, and vice versa.

$\textbf{M}^1$, $\textbf{M}^2$, and $\textbf{M}^3$, the set of pairwise cross-modal similarity matrices, represent the similarity among images, keyframes and videos. The probability of similarity between  image $\mathbf{x}_i$ and video $\mathbf{z}_k$ given the corresponding domain-invariant feature vectors $\mathbf{F}_{*i}$ and $\mathbf{G}_{*k}$, can be expressed as a likelihood function defined as follows:
\begin{equation}\label{eq1}
\begin{split}
  p(M^1_{ik}|\mathbf{F}_{*i},\mathbf{G}_{*k})&=      \left\{
          \begin{array}{ll}
            \sigma(\Theta^1_{ik}) & \hbox{$M^1_{ik}=1$} \\
            1-\sigma(\Theta^1_{ik}) & \hbox{$M^1_{ik}=0$}
          \end{array}
        \right.\\
&=\sigma(\Theta^1_{ik}) ^{M^1_{ik}}(1-\sigma(\Theta^1_{ik}))^{1-M^1_{ik}}
\end{split}
\end{equation}
where $\Theta^1_{ik}=\frac{1}{2}\mathbf{F}_{*i}^{\textrm{T}}\mathbf{G}_{*k}$ is the inner product between domain-invariant feature vectors for image $\mathbf{x}_i$ and video $\mathbf{z}_k$, and $\sigma(\Theta^1_{ik})=\frac{1}{1+e^{-\Theta^1_{ik}}}$ is the sigmoid function. As the inner product $\Theta^1_{ik}$ increases, the probability of $p(M^1_{ik}=1|\mathbf{F}_{*i},\mathbf{G}_{*k})$ also increases, i.e., $\mathbf{x}_i$ and $\mathbf{z}_k$ belong to the same action class. As the inner product $\Theta^1_{ik}$ decreases, the probability of $p(M^1_{ik}=1|\mathbf{F}_{*i},\mathbf{G}_{*k})$ also decreases, i.e., $\mathbf{x}_i$ and $\mathbf{z}_k$ belong to different action classes.

Similarly, the probability of similarity between image $\mathbf{x}_i$ and keyframe $\mathbf{y}_j$, and the probability of similarity between keyframe $\mathbf{y}_j$ and video $\mathbf{z}_k$ can be expressed as two likelihood functions defined as follows:
\begin{equation}\label{eq2}
\begin{split}
  p(M^2_{ij}|\mathbf{F}_{*i},\mathbf{H}_{*j})&=      \left\{
          \begin{array}{ll}
            \sigma(\Theta^2_{ij}) & \hbox{$M^2_{ij}=1$} \\
            1-\sigma(\Theta^2_{ij}) & \hbox{$M^2_{ij}=0$}
          \end{array}
        \right.\\
&=\sigma(\Theta^2_{ij}) ^{M^2_{ij}}(1-\sigma(\Theta^2_{ij}))^{1-M^2_{ij}}
\end{split}
\end{equation}
\begin{equation}\label{eq3}
\begin{split}
  p(M^3_{jk}|\mathbf{H}_{*j},\mathbf{G}_{*k})&=      \left\{
          \begin{array}{ll}
            \sigma(\Theta^3_{jk}) & \hbox{$M^3_{jk}=1$} \\
            1-\sigma(\Theta^3_{jk}) & \hbox{$M^3_{jk}=0$}
          \end{array}
        \right.\\
&=\sigma(\Theta^3_{jk}) ^{M^3_{jk}}(1-\sigma(\Theta^3_{jk}))^{1-M^3_{jk}}
\end{split}
\end{equation}

Therefore, the negative log likelihood of the similarity matrix $\mathbf{M^1}$ given $\mathbf{F}$ and $\mathbf{G}$ can be written as follows:
\begin{equation}\label{eq4}
\begin{split}
\mathcal{L}(\mathbf{F},\mathbf{G})&=-\textrm{log}~p(\mathbf{M^1}|\mathbf{F},\mathbf{G})~\\&=-\mathop{\sum}_{i,k=1}^{n}((M^1_{ik}-1)\Theta^1_{ik}-\textrm{log}(1+\textrm{exp}(\Theta^1_{ik})))
\end{split}
\end{equation}

Similarly, the negative log likelihood of the similarity matrix $\mathbf{M^2}$ given $\mathbf{F}$ and $\mathbf{H}$ can be written as follows:
\begin{equation}\label{eq5}
\begin{split}
\mathcal{L}(\mathbf{F},\mathbf{H})&=-\textrm{log}~p(\mathbf{M^2}|\mathbf{F},\mathbf{H})~\\&=-\mathop{\sum}_{i,j=1}^{n}((M^2_{ij}-1)\Theta^2_{ij}-\textrm{log}(1+\textrm{exp}(\Theta^2_{ij})))
\end{split}
\end{equation}

The negative log likelihood of the similarity matrix $\mathbf{M^3}$ given $\mathbf{H}$ and $\mathbf{G}$ can also be written as follows:
\begin{equation}\label{eq6}
\begin{split}
\mathcal{L}(\mathbf{H},\mathbf{G})&=-\textrm{log}~p(\mathbf{M^3}|\mathbf{H},\mathbf{G})~\\&=-\mathop{\sum}_{j,k=1}^{n}((M^3_{jk}-1)\Theta^3_{jk}-\textrm{log}(1+\textrm{exp}(\Theta^3_{jk})))
\end{split}
\end{equation}
It is easy to find that minimizing these negative log likelihood is equivalent to maximizing the likelihood, which can simultaneously encourage the similarity (inner product) between $\mathbf{F}_{*i}$ and $\mathbf{G}_{*k}$, the similarity (inner product) between $\mathbf{F}_{*i}$ and $\mathbf{H}_{*j}$, and the similarity (inner product) between $\mathbf{H}_{*j}$ and $\mathbf{G}_{*k}$ to be large when $M^1_{ik}=1$, $M^2_{ij}=1$ and $M^3_{jk}=1$, and to be small when $M^1_{ik}=0$, $M^2_{ij}=0$ and $M^3_{jk}=0$. Therefore, optimizing these cross-modal similarities metric terms in Eqs. (\ref{eq4})-(\ref{eq6}) can preserve the cross-modal similarities in $M^1$, $M^2$ and $M^3$, and learn domain-invariant representations for image, keyframe and video modalities.

Now we show how to utilize the semantic relationship among images, keyframes and videos to effectively fuse the learned domain-invariant keyframe and video representations. We adopt an autoencoder architecture, whose
hidden layer is constrained to be equal to the semantic representation of the names of the actions (e.g. attribute representations \cite{DBLP:conf/cvpr/LampertNH09} or word2vec representations \cite{DBLP:conf/nips/MikolovSCCD13}). To realize this, we force the latent space $\mathbf{S}$ to be the $k$-dimensional semantic representation space, e.g. each column of $\mathbf{S}$ is an attribute or word2vec vector of an action class name. Specifically, given learned domain-invariant image representation matrix $\mathbf{F}\in\mathbb{R}^{d\times n}$, keyframe representation matrix $\mathbf{H}\in\mathbb{R}^{d\times n}$
and video representation matrix $\mathbf{G}\in\mathbb{R}^{d\times n}$ as the inputs, we construct three semantic autoencoders named Auto-F, Auto-H and Auto-G, for image modality, keyframe modality and video modality, respectively. The objective functions are defined as follows:
\begin{equation}\label{eq7}
  \mathop{\textrm{min}}_{\mathbf{W}_F,\mathbf{W}_{F}^\ast}\|\mathbf{F}-\mathbf{W}_{F}^\ast\mathbf{W}_{F}\mathbf{F}\|_{F}^{2}~~~~s.t.~~~ \mathbf{W}_{F}\mathbf{F}=\mathbf{S}_i
\end{equation}
\begin{equation}\label{eq8}
  \mathop{\textrm{min}}_{\mathbf{W}_H,\mathbf{W}_{H}^\ast}\|\mathbf{H}-\mathbf{W}_{H}^\ast\mathbf{W}_{H}\mathbf{H}\|_{F}^{2}~~~~s.t.~~~ \mathbf{W}_{H}\mathbf{H}=\mathbf{S}_f
\end{equation}
\begin{equation}\label{eq9}
  \mathop{\textrm{min}}_{\mathbf{W}_G,\mathbf{W}_{G}^\ast}\|\mathbf{G}-\mathbf{W}_{G}^\ast\mathbf{W}_{G}\mathbf{G}\|_{F}^{2}~~~~s.t.~~~ \mathbf{W}_{G}\mathbf{G}=\mathbf{S}_v
\end{equation}
where $\mathbf{W}_F\in\mathbb{R}^{k\times d}$,  and $\mathbf{W}_{F}^\ast\in\mathbb{R}^{d\times k}$ are projection matrices for Auto-F, $\mathbf{W}_H\in\mathbb{R}^{k\times d}$ and $\mathbf{W}_{H}^\ast\in\mathbb{R}^{d\times k}$ are projection matrices for Auto-H, $\mathbf{W}_G\in\mathbb{R}^{k\times d}$ and $\mathbf{W}_{G}^\ast\in\mathbb{R}^{d\times k}$ are projection matrices for Auto-G, $\mathbf{S}_i\in \mathbb{R}^{k\times n}$, $\mathbf{S}_f\in \mathbb{R}^{k\times n}$ and $\mathbf{S}_v\in \mathbb{R}^{k\times n}$ are the semantic representations for image modality, keyframe modality and video modality, respectively. The feature representations from the hidden layer of these three semantic autoencoders contain both learned domain-invariant knowledge and semantic knowledge. In this way, more
informative and discriminative features can be obtained.

Since the semantic representations of the same action classes from images and videos are the same, which means that $\mathbf{S}_i=\mathbf{S}_f=\mathbf{S}_v$. Thus, we uniformly use $\mathbf{S}$ to denote the semantic representations for these modalities. To further simplify the objective functions  Eqs. (\ref{eq7})-(\ref{eq9}), we consider tied weights \cite{DBLP:conf/nips/RanzatoBL07} such that $\mathbf{W}^\ast=\mathbf{W}^{\top}$ and substitute $\mathbf{W}_{F}\mathbf{F}$,  $\mathbf{W}_{H}\mathbf{H}$ and $\mathbf{W}_{G}\mathbf{G}$ with $\mathbf{S}$. In addition, we relax the constraint into a soft one since hard constraint such as $\mathbf{W}_{F}\mathbf{F}=\mathbf{S}$ is difficult to solve. Finally, the objective functions Eqs. (\ref{eq7})-(\ref{eq9}) can be rewritten as follows:
\begin{equation}\label{eq10}
  \mathop{\textrm{min}}_{\mathbf{W}_F}\|\mathbf{F}-\mathbf{W}_{F}^{\top}\mathbf{S}\|_{F}^{2}+\lambda\|\mathbf{W}_F\mathbf{F}-\mathbf{S}\|_{F}^{2}
\end{equation}
\begin{equation}\label{eq11}
  \mathop{\textrm{min}}_{\mathbf{W}_H}\|\mathbf{H}-\mathbf{W}_{H}^{\top}\mathbf{S}\|_{F}^{2}+\lambda\|\mathbf{W}_H\mathbf{H}-\mathbf{S}\|_{F}^{2}
\end{equation}
\begin{equation}\label{eq12}
  \mathop{\textrm{min}}_{\mathbf{W}_G}\|\mathbf{G}-\mathbf{W}_{G}^{\top}\mathbf{S}\|_{F}^{2}+\lambda\|\mathbf{W}_G\mathbf{G}-\mathbf{S}\|_{F}^{2}
\end{equation}
where $\lambda$ is a weighting coefficient controlling the importance of the decoder and encoder.

Since the modality gap among images, video keyframes and videos has been reduced, the learned domain-invariant keyframe feature contains knowledge from both image and keyframe modalities. Thus, we can enhance action recognition performance in videos by fusing the learned domain-invariant keyframe features and video features in a simple yet effective way. Specifically, we construct the fourth semantic autoencoder named Auto-E, with the concatenation of learned domain-invariant keyframe representations $\mathbf{H}\in\mathbb{R}^{d\times n}$ and video representations $\mathbf{G}\in\mathbb{R}^{d\times n}$ as input, which is denoted as $\mathbf{E}=[\mathbf{H};\mathbf{G}]\in\mathbb{R}^{2d\times n}$. The objective function of this semantic autoencoder is defined as follows:
\begin{equation}\label{eq13}
  \mathop{\textrm{min}}_{\mathbf{W}_E}\|\mathbf{E}-\mathbf{W}_{E}^{\top}\mathbf{S}\|_{F}^{2}+\lambda\|\mathbf{W}_E\mathbf{E}-\mathbf{S}\|_{F}^{2}
\end{equation}
where $\mathbf{W}_E\in\mathbb{R}^{k\times 2d}$  is the projection matrix for Auto-E. In this way, the hidden unit is a good fused representation for both keyframe and video modalities, as the reconstruction process captures the intrinsic structure of the domain-invariant space and the semantic relationship between keyfame and videos.

To this end, we integrate cross-modal feature fusion Eqs. (\ref{eq10})-(\ref{eq13}) and domain-invariant representations learning Eqs. (\ref{eq4})-(\ref{eq6}) into a unified deep learning framework named Deep Image-to-Video Adaptation and Fusion Networks (DIVAFN), and the final objective function is defined as follows:
\begin{equation}\label{eq14}
\begin{split}
&\mathop{\textrm{min}}_{\theta_x,\theta_y,\theta_z, \mathbf{W}_F,\mathbf{W}_H,\mathbf{W}_G,\mathbf{W}_E}\mathcal{J}=~\\
&-a\mathop{\sum}_{i,k=1}^{n}((M^1_{ik}-1)\Theta^1_{ik}-\textrm{log}(1+\textrm{exp}(\Theta^1_{ik})))~\\
&-b\mathop{\sum}_{i,j=1}^{n}((M^2_{ij}-1)\Theta^2_{ij}-\textrm{log}(1+\textrm{exp}(\Theta^2_{ij})))~\\
&-c\mathop{\sum}_{j,k=1}^{n}((M^3_{jk}-1)\Theta^3_{jk}-\textrm{log}(1+\textrm{exp}(\Theta^3_{jk})))~\\
&+\beta(\|\mathbf{F}-\mathbf{W}_{F}^{\top}\mathbf{S}\|_{F}^{2}+\|\mathbf{H}-\mathbf{W}_{H}^{\top}\mathbf{S}\|_{F}^{2}+\|\mathbf{G}-\mathbf{W}_{G}^{\top}\mathbf{S}\|_{F}^{2}~\\
&+\|\mathbf{E}-\mathbf{W}_{E}^{\top}\mathbf{S}\|_{F}^{2})+\lambda(\|\mathbf{W}_F\mathbf{F}-\mathbf{S}\|_{F}^{2}+\|\mathbf{W}_H\mathbf{H}-\mathbf{S}\|_{F}^{2}~\\
&+\|\mathbf{W}_G\mathbf{G}-\mathbf{S}\|_{F}^{2}+\|\mathbf{W}_E\mathbf{E}-\mathbf{S}\|_{F}^{2})
\end{split}
\end{equation}
where $a$, $b$ and $c$ denote the weighting coefficients controlling the importance of the three negative log likelihood functions, $\beta$ and $\lambda$ are weighting coefficients controlling the importance of the decoders and encoders, respectively. To make the objective function concise and converge more easily, the values of $\beta$ and $\lambda$ for all semantic autoencoders are encouraged to be the same. The network parameters of the CNN for image, keyframe and video modalities are defined as $\theta_x$, $\theta_y$  and $\theta_z$, respectively. Since there are multiple variables in Eq. (\ref{eq14}), we propose an efficient alternating optimization algorithm for DIVAFN.
\subsection{Optimization}
We adopt an alternating learning strategy to learn $\theta_x$, $\theta_y$, $\theta_z$, $\mathbf{W}_F$, $\mathbf{W}_H$, $\mathbf{W}_G$ and $\mathbf{W}_E$. Each time we learn one variable with other variables fixed.
\subsubsection{Learn $\theta_x$, with $\theta_y$, $\theta_z$, $\mathbf{W}_F$, $\mathbf{W}_H$, $\mathbf{W}_G$ and $\mathbf{W}_E$ fixed}
When $\theta_y$, $\theta_z$, $\mathbf{W}_F$, $\mathbf{W}_H$, $\mathbf{W}_G$ and $\mathbf{W}_E$ are fixed, we learn the CNN parameter $\theta_x$ of the image modality by back-propagation (BP) algorithm. As the most existing deep learning approaches \cite{DBLP:conf/nips/KrizhevskySH12}, we adopt the Stochastic Gradient Descent (SGD) to learn $\theta_x$ with the BP algorithm. In each iteration, a mini-batch of samples from the training set is used to learn the variable. Specifically, for each training sample $\mathbf{x}_i$ from image modality, we first compute the following gradient:
\begin{equation}\label{eq15}
\begin{split}
\frac{\partial\mathcal{J}}{\partial \mathbf{F}_{*i}}&=\frac{a}{2}\mathop{\sum}_{k=1}^{n}(\sigma(\Theta^1_{ik})\mathbf{G}_{*k}-(M^1_{ik}-1)\mathbf{G}_{*k})~\\
&+\frac{b}{2}\mathop{\sum}_{j=1}^{n}(\sigma(\Theta^2_{ij})\mathbf{H}_{*j}-(M^2_{ij}-1)\mathbf{H}_{*j})~\\
&+2\beta(\mathbf{F}_{*i}-\mathbf{W}_F^{\top}\mathbf{S}_{*i})+2\lambda \mathbf{W}_F^{\top}(\mathbf{W}_F\mathbf{F}_{*i}-\mathbf{S}_{*i})
\end{split}
\end{equation}
Then we can compute $\frac{\partial\mathcal{J}}{\partial \theta_x}$ with $\frac{\partial\mathcal{J}}{\partial \mathbf{F}_{*i}}$ using the chain rule, based on which the BP algorithm can be used to update the variable $\theta_x$.
\subsubsection{Learn $\theta_y$, with $\theta_x$, $\theta_z$, $\mathbf{W}_F$, $\mathbf{W}_H$, $\mathbf{W}_G$ and $\mathbf{W}_E$ fixed}
When $\theta_x$, $\theta_z$, $\mathbf{W}_F$, $\mathbf{W}_H$, $\mathbf{W}_G$ and $\mathbf{W}_E$ are fixed, we also learn the CNN parameter $\theta_y$ of the keyframe modality by using SGD with back-propagation (BP) algorithm. Specifically, for each training sample $\mathbf{y}_j$ from keyframe modality, we first compute the following gradient:
\begin{equation}\label{eq16}
\begin{split}
\frac{\partial\mathcal{J}}{\partial \mathbf{H}_{*j}}&=\frac{b}{2}\mathop{\sum}_{j=1}^{n}(\sigma(\Theta^2_{ij})\mathbf{F}_{*j}-(M^2_{ij}-1)\mathbf{F}_{*j})~\\
&+\frac{c}{2}\mathop{\sum}_{k=1}^{n}(\sigma(\Theta^3_{jk})\mathbf{G}_{*k}-(M^3_{jk}-1)\mathbf{G}_{*k})~\\
&+2\beta(\mathbf{H}_{*j}-\mathbf{W}_H^{\top}\mathbf{S}_{*j})+2\lambda \mathbf{W}_H^{\top}(\mathbf{W}_H\mathbf{H}_{*j}-\mathbf{S}_{*j})
\end{split}
\end{equation}

Then we can compute $\frac{\partial\mathcal{J}}{\partial \theta_y}$ with $\frac{\partial\mathcal{J}}{\partial \mathbf{H}_{*j}}$ using the chain rule, based on which the BP algorithm can be used to update the variable $\theta_y$.

\subsubsection{Learn $\theta_z$, with $\theta_x$, $\theta_y$, $\mathbf{W}_F$, $\mathbf{W}_H$, $\mathbf{W}_G$ and $\mathbf{W}_E$ fixed}
When $\theta_x$, $\theta_y$, $\mathbf{W}_F$, $\mathbf{W}_H$, $\mathbf{W}_G$ and $\mathbf{W}_E$ are fixed, we also learn the CNN parameter $\theta_z$ of the video modality by using SGD with back-propagation (BP) algorithm. Specifically, for each training sample $\mathbf{z}_k$ from video modality, we first compute the following gradient:
\begin{equation}\label{eq17}
\begin{split}
\frac{\partial\mathcal{J}}{\partial \mathbf{G}_{*k}}&=\frac{a}{2}\mathop{\sum}_{i=1}^{n}(\sigma(\Theta^1_{ki})\mathbf{F}_{*i}-(M^1_{ki}-1)\mathbf{F}_{*i})~\\
&+\frac{c}{2}\mathop{\sum}_{j=1}^{n}(\sigma(\Theta^3_{kj})\mathbf{H}_{*j}-(M^3_{kj}-1)\mathbf{H}_{*j})~\\
&+2\beta(\mathbf{G}_{*k}-\mathbf{W}_G^{\top}\mathbf{S}_{*k})+2\lambda \mathbf{W}_G^{\top}(\mathbf{W}_G\mathbf{G}_{*k}-\mathbf{S}_{*k})
\end{split}
\end{equation}

Then we can compute $\frac{\partial\mathcal{J}}{\partial \theta_z}$ with $\frac{\partial\mathcal{J}}{\partial \mathbf{G}_{*k}}$ using the chain rule, based on which the BP algorithm can be used to update the variable $\theta_z$.

\subsubsection{Learn $\mathbf{W}_F$, with $\theta_x$, $\theta_y$, $\theta_z$, $\mathbf{W}_G$, $\mathbf{W}_H$ and $\mathbf{W}_E$ fixed}
When $\theta_x$, $\theta_y$, $\theta_z$, $\mathbf{W}_G$,  $\mathbf{W}_H$ and $\mathbf{W}_E$ are fixed, the objective function Eq. (\ref{eq14}) can be rewritten as follows:
\begin{equation}\label{eq18}
  \mathop{\textrm{min}}_{\mathbf{W}_F}\beta\|\mathbf{F}-\mathbf{W}_{F}^{\top}\mathbf{S}\|_{F}^{2}+\lambda\|\mathbf{W}_F\mathbf{F}-\mathbf{S}\|_{F}^{2}
\end{equation}
Since Eq. (\ref{eq18}) is quadratic, it is a convex function with a global optimal solution. To optimize it, we first reorganize Eq. (\ref{eq18}) using trace properties: $\textrm{Tr}(\mathbf{X}^{\top})=\textrm{Tr}(\mathbf{X})$ and $\textrm{Tr}(\mathbf{S}^{\top}\mathbf{W})=\textrm{Tr}(\mathbf{W}^{\top}\mathbf{S})$:
\begin{equation}\label{eq19}
  \mathop{\textrm{min}}_{\mathbf{W}_F}\beta\|\mathbf{F}^{\top}-\mathbf{S}^{\top}\mathbf{W}_{F}\|_{F}^{2}+\lambda\|\mathbf{W}_F\mathbf{F}-\mathbf{S}\|_{F}^{2}
\end{equation}
And then we set the derivative of Eq. (\ref{eq19}) with respect to $\mathbf{W}_F$ to zero and get:
\begin{equation}\label{eq20}
\begin{split}
 -\beta \mathbf{S}(\mathbf{F}^{\top}-\mathbf{S}^{\top}\mathbf{W}_F)+\lambda(\mathbf{W}_F\mathbf{F}-\mathbf{S})\mathbf{F}^{\top}=0~\\
\beta \mathbf{S} \mathbf{S}^{\top}\mathbf{W}_F+\lambda\mathbf{W}_F\mathbf{F}\mathbf{F}^{\top}=\beta\mathbf{S}\mathbf{F}^{\top}+\lambda\mathbf{S}\mathbf{F}^{\top}
\end{split}
\end{equation}
If we denote $\mathbf{A}=\beta \mathbf{S} \mathbf{S}^{\top}$, $\mathbf{B}=\lambda\mathbf{F}\mathbf{F}^{\top}$, and $\mathbf{C}=(\beta+\lambda)\mathbf{S}\mathbf{F}^{\top}$, we have the following well-known Sylvester formulation:
\begin{equation}\label{eq21}
\mathbf{A}\mathbf{W}_F+\mathbf{W}_F\mathbf{B}=\mathbf{C}
\end{equation}
which can be solved efficiently by the Bartels-Stewart algorithm \cite{DBLP:journals/cacm/BartelsS72}. More importantly, the calculation complexity of Eq. (\ref{eq21}) only depends on the feature dimension ($\mathcal{O}(d^3)$) and thus it can scale to large-scale datasets.
\subsubsection{Learn $\mathbf{W}_H$, with $\theta_x$, $\theta_y$, $\theta_z$, $\mathbf{W}_F$,  $\mathbf{W}_G$ and $\mathbf{W}_E$ fixed}
When $\theta_x$, $\theta_y$, $\theta_z$,  $\mathbf{W}_F$, $\mathbf{W}_G$ and $\mathbf{W}_E$ are fixed, the objective function Eq. (\ref{eq14}) can be rewritten as follows:
\begin{equation}\label{eq22}
  \mathop{\textrm{min}}_{\mathbf{W}_H}\beta\|\mathbf{H}-\mathbf{W}_{H}^{\top}\mathbf{S}\|_{F}^{2}+\lambda\|\mathbf{W}_H\mathbf{H}-\mathbf{S}\|_{F}^{2}
\end{equation}
Similarly, Eq. (\ref{eq22}) can also be solved efficiently by the Bartels-Stewart algorithm \cite{DBLP:journals/cacm/BartelsS72}.
\subsubsection{Learn $\mathbf{W}_G$, with $\theta_x$, $\theta_y$, $\theta_z$, $\mathbf{W}_F$, $\mathbf{W}_H$ and $\mathbf{W}_E$ fixed}
When $\theta_x$, $\theta_y$, $\theta_z$, $\mathbf{W}_F$, $\mathbf{W}_H$ and $\mathbf{W}_E$ are fixed, the objective function Eq. (\ref{eq14}) can be rewritten as follows:
\begin{equation}\label{eq23}
  \mathop{\textrm{min}}_{\mathbf{W}_G}\beta\|\mathbf{G}-\mathbf{W}_{G}^{\top}\mathbf{S}\|_{F}^{2}+\lambda\|\mathbf{W}_G\mathbf{G}-\mathbf{S}\|_{F}^{2}
\end{equation}
Similarly, Eq. (\ref{eq23}) can also be solved efficiently by the Bartels-Stewart algorithm \cite{DBLP:journals/cacm/BartelsS72}.
\subsubsection{Learn $\mathbf{W}_E$, with $\theta_x$, $\theta_y$, $\theta_z$, $\mathbf{W}_F$, $\mathbf{W}_H$ and $\mathbf{W}_G$ fixed}
When $\theta_x$, $\theta_y$, $\theta_z$, $\mathbf{W}_F$, $\mathbf{W}_H$ and $\mathbf{W}_G$ are fixed, the objective function Eq. (\ref{eq14}) can be rewritten as follows:
\begin{equation}\label{eq24}
  \mathop{\textrm{min}}_{\mathbf{W}_E}\beta\|\mathbf{E}-\mathbf{W}_{E}^{\top}\mathbf{S}\|_{F}^{2}+\lambda\|\mathbf{W}_E\mathbf{E}-\mathbf{S}\|_{F}^{2}
\end{equation}
Similarly, Eq. (\ref{eq24}) can also be solved efficiently by the Bartels-Stewart algorithm \cite{DBLP:journals/cacm/BartelsS72}.

\subsection{Classification}
When the loss of the objective function Eq. (\ref{eq14}) drops to its minimum value after a certain number of iterations, we can learn variables $\theta_x$, $\theta_y$, $\theta_z$, $\mathbf{W}_F$, $\mathbf{W}_H$, $\mathbf{W}_G$ and $\mathbf{W}_E$. Then we use the concatenation of the learned semantic feature representations from the hidden layers of Auto-H, Auto-G and Auto-E as the final fused feature representations. For example, given the video sample $\mathbf{z}_k$, the final fused feature representation for action recognition in videos is defined as:
\begin{equation}\label{eq25}
\begin{split}
&\mathbf{B}_{*k}=[\hat{\mathbf{S}}_f^k; \hat{\mathbf{S}}_v^k; \hat{\mathbf{S}}^k]~\\
&=[\mathbf{W}_Hf(\mathbf{y}_j;\theta_y);\mathbf{W}_Gg(\mathbf{z}_k;\theta_z);\mathbf{W}_E[f(\mathbf{y}_j;\theta_y);g(\mathbf{z}_k;\theta_z)]]
\end{split}
\end{equation}
where $[ \cdot ; \cdot ]$ denotes the matrix concatenation operation, $\hat{\mathbf{S}}_f^k\in \mathbb{R}^{k}$ denotes the learned keyframe semantic representation from Auto-H for video sample $\mathbf{z}_k$, $\hat{\mathbf{S}}_v^k\in \mathbb{R}^{k}$ denotes the learned video semantic representation from Auto-G for video sample $\mathbf{z}_k$, $\hat{\mathbf{S}}^k\in \mathbb{R}^{k}$ denotes the fused semantic representation from Auto-E, and $\mathbf{B}_{*k}\in \mathbb{R}^{3k}$ denotes the finally fused feature representation for video sample $\mathbf{z}_k$. Finally, we use the final fused representation matrix $\mathbf{B}\in \mathbb{R}^{3k\times n}$ to perform action recognition in videos using the SVM \cite{DBLP:journals/tist/ChangL11} classifier, where $n$ denotes the number of training samples.

\section{Experiments}

\subsection{Datasets}
\begin{figure}[!t]
\centering
\includegraphics[scale=0.235]{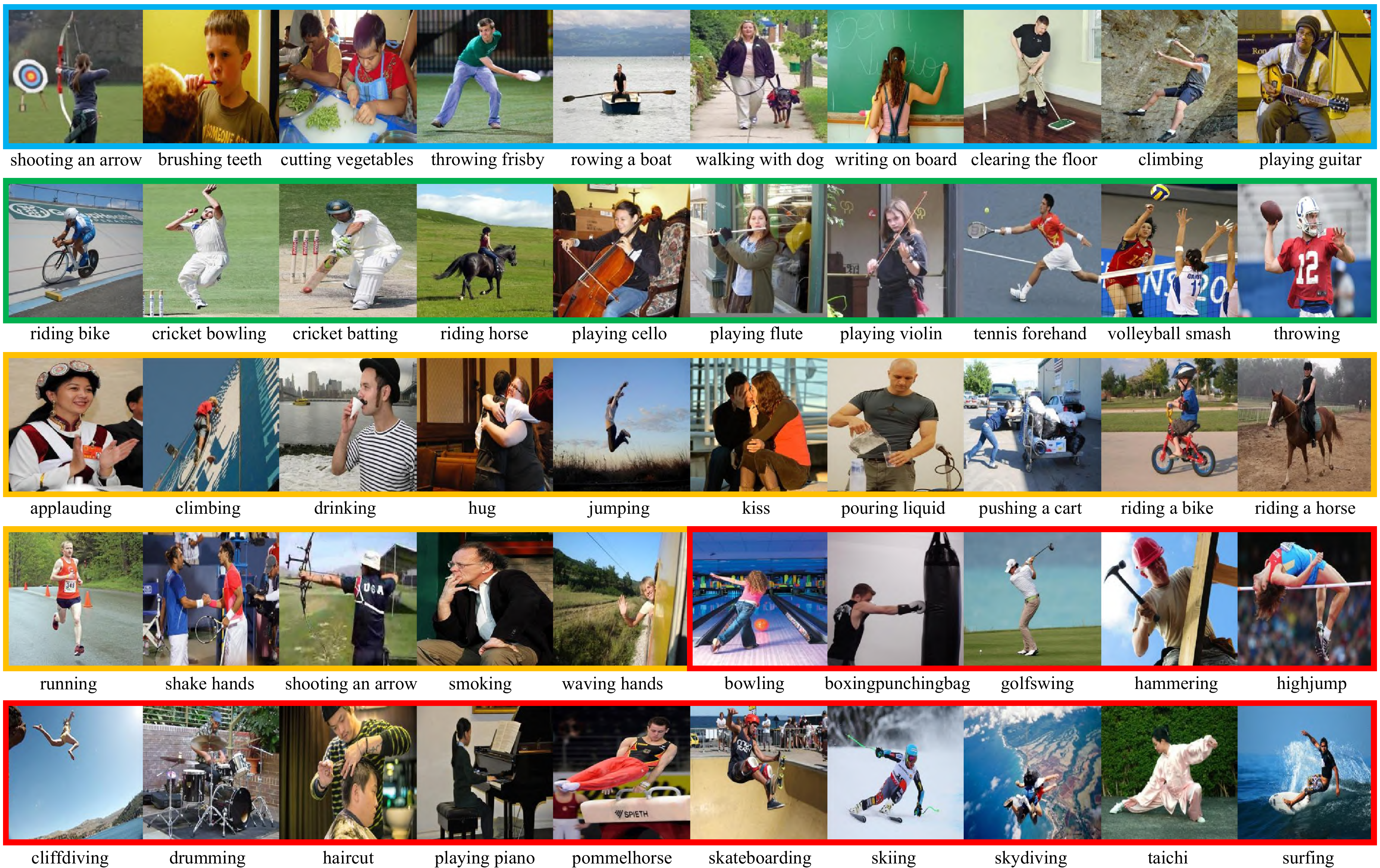}
\caption{Sample images from each image dataset, where the images with bounding boxes in blue, green, orange and red colors are from the image dataset of Stanford40$\rightarrow$UCF101, ASD$\rightarrow$UF101, EAD$\rightarrow$HMDB51 and BU101$\rightarrow$UCF101 datasets, respectively.}
\label{Fig3}
\end{figure}

In order to evaluate the performance of our method, we conduct experiments on four real-world image-video action recognition datasets. Among these datasets, videos come from two large-scale and complex datasets, i.e., HMDB51 \cite{DBLP:conf/iccv/KuehneJGPS11} and UCF101 \cite{DBLP:journals/corr/abs-1212-0402}. And images come from various datasets, e.g. BU101 dataset \cite{DBLP:journals/pr/MaBZSS17}, Stanford40 dataset \cite{DBLP:conf/iccv/YaoJKLGF11}, Action dataset (DB) \cite{DBLP:journals/pami/GuptaKD09}, people playing musical instrument (PPMI) dataset \cite{DBLP:conf/cvpr/YaoF10}, Willow-Actions dataset \cite{DBLP:conf/bmvc/DelaitreLS10}, Still DB \cite{DBLP:conf/icpr/IkizlerCPD08}, and Human Interaction Image (HII) dataset \cite{DBLP:conf/mm/LiWZK17}. Figure \ref{Fig3} shows the sample images from each dataset.

\subsubsection{Stanford40$\rightarrow$UCF101 (S$\rightarrow$U)}

The UCF101 \cite{DBLP:journals/corr/abs-1212-0402} is a video action dataset collected from YouTube with $101$ action classes. The Stanford40 dataset \cite{DBLP:conf/iccv/YaoJKLGF11} contains $40$ diverse human action images with various pose, appearance and background clutter. We choose $10$ common classes between these two datasets, and each action class has $110$ samples for both image and video modalities. In Table \ref{Table 1}, we summarize the chosen common classes of Stanford40$\rightarrow$UCF101, where ``$\rightarrow$'' denotes the direction of adaptation from auxiliary modality to target modality.

\begin{table}[!t]\renewcommand\tabcolsep{3pt}
            \renewcommand{\arraystretch}{1.0}
              \caption{Chosen common classes of Stanford40$\rightarrow$UCF101.}
              \label{Table 1}
            \centering
            \begin{tabular}{|c|c|c|}\hline
            Datasets&Stanford40&UCF101\\\hline
            \multirow{10}{*}{Common classes}&shooting an arrow&archery\\
            &brushing teeth&brushing teeth\\
            &cutting vegetables&cutting in kitchen\\
            &throwing frisby&frisbee catch\\
            &rowing a boat&rowing\\
            &walking with dog&walking with dog\\
            &writing on board&writing on board\\
            &clearing the floor&mopping floor\\
            &climbing&rock climbing indoor\\
            &playing guitar&playing guitar\\\hline
            \end{tabular}
 \end{table}

 \begin{table}[!t]\renewcommand\tabcolsep{3pt}
                \renewcommand{\arraystretch}{1.0}
              \caption{Chosen common classes of ASD$\rightarrow$UCF101. }
              \label{Table 2}
                \centering
                \begin{tabular}{|c|c|c|}\hline
                Datasets&ASD (source dataset)&UCF101\\\hline
                \multirow{10}{*}{Common classes}&riding bike (Willow Actions)&biking\\
                &cricket bowling (Action DB)&cricket bowling\\
                &cricket batting (Action DB)&cricket shot\\
                &riding horse (Willow Actions)&horse riding\\
                &playing cello (PPMI)&playing cello\\
                &playing flute (PPMI)&playing flute\\
                &playing violin (PPMI)&playing violin\\
                &tennis forehand (Action DB)&tennis swing\\
                &volleyball smash (Action DB)&volleyball spiking\\
                &throwing (Still DB)&baseball pitch\\\hline
                 \end{tabular}
  \end{table}

 \begin{table}[!t]\renewcommand\tabcolsep{3pt}
                \renewcommand{\arraystretch}{1.0}
              \caption{Chosen common classes of EAD$\rightarrow$HMDB51. }
              \label{Table 3}
                \centering
                \begin{tabular}{|c|c|c|}\hline
                Datasets&EAD (source dataset)&HMDB51\\\hline
                \multirow{15}{*}{Common classes} &applauding (Stanford40)&clap\\
                &climbing (Stanford40)&climb\\
                &drinking (Stanford40)&drink\\
                &hug (HII)&hug\\
                &jumping (Stanford40)&jump\\
                &kiss (HII)&kiss\\
                &pouring liquid (Stanford40)&pour\\
                &pushing a cart (Stanford40)&push\\
                &riding a bike (Stanford40)&ride bike\\
                &riding a horse (Stanford40)&ride horse\\
                &running (Stanford40)&run\\
                &shake hands (HII)&shake hands\\
                &shooting an arrow (Stanford40)&shoot bow\\
                &smoking (Stanford40)&smoke\\
                &waving hands (Stanford40)&wave\\\hline
                 \end{tabular}
  \end{table}

\subsubsection{ASD$\rightarrow$UCF101 (A$\rightarrow$U)}
To further evaluate whether images coming from different sources may influence the action recognition performance in videos, we select $10$ common classes between UCF101 and an extended dataset named ``actions from still datasets'' (ASD), which consists of four publicly available datasets, i.e., Action dataset (DB) \cite{DBLP:journals/pami/GuptaKD09}, people playing musical instrument (PPMI) dataset \cite{DBLP:conf/cvpr/YaoF10}, Willow-Actions dataset \cite{DBLP:conf/bmvc/DelaitreLS10}, and Still DB \cite{DBLP:conf/icpr/IkizlerCPD08}. Each action class has $50$ samples for both image and video modalities. Table \ref{Table 2} shows the chosen common classes of ASD$\rightarrow$UCF101.

\subsubsection{EAD$\rightarrow$HMDB51 (E$\rightarrow$H)}
The HMDB51 \cite{DBLP:conf/iccv/KuehneJGPS11} dataset consists of $51$ classes with a total of $6766$ video clips. We construct an ``extensive action dataset'' (EAD) which consists of the chosen categories from Stanford40 \cite{DBLP:conf/iccv/YaoJKLGF11} and HII datasets \cite{DBLP:conf/mm/LiWZK17}. The HII dataset contains web action images from search engines (Google, Bing and Flickr) with four types of interactions: handshake, highfive, hug, and kiss. We choose $15$ common classes and each class has $100$ samples for both image and video modalities. Table \ref{Table 3} shows the chosen common classes of EAD$\rightarrow$HMDB51.
\subsubsection{BU101$\rightarrow$UCF101 (B$\rightarrow$U)}
To investigate the performance of our method for large-scale dataset, we select BU101 \cite{DBLP:journals/pr/MaBZSS17} as the image dataset, which is the largest web action image dataset until now. This dataset is more than double the size of the largest previous action image dataset Stanford40 \cite{DBLP:conf/iccv/YaoJKLGF11}, both in the number of images and the number of actions. It consists of $23.8$K action images and has one-to-one correspondence to the $101$ action classes in the UCF101 video dataset. For video dataset, we select all the $13320$ videos with $101$ classes from the UCF101 dataset. Therefore, we can evaluate the action recognition performance in UCF101 using the public three train/test splits and make a fair comparison with other state-of-the-art action recognition algorithms.

\subsection{Experimental Setup}
To study the performance variance when the ratios of training samples are different, we select various amount of training samples to evaluate our method. For Stanford40$\rightarrow$UCF101, ASD$\rightarrow$UCF101 and EAD$\rightarrow$HMDB51 datasets,  the ratios of training samples are set to $10\%$, $20\%$, $30\%$, $40\%$ and $50\%$. For BU101$\rightarrow$ UCF101 dataset, we use the public three train/test splits and randomly sample $5\%$, $10\%$, $20\%$, $33\%$, $50\%$ and $100\%$ videos from the training set of each split as training samples for our method, and the result is averaged and reported across three train/test splits.

We exploit the CNN-F network \cite{DBLP:conf/bmvc/ChatfieldSVZ14} pre-trained on ImageNet dataset \cite{DBLP:journals/ijcv/RussakovskyDSKS15} to initialize the first seven layers of the CNN for image and keyframe modalities. All the other parameters of the deep neural networks in DIVAFN are randomly initialized. For image and keyframe modalities, we use the raw image pixels as the input to the deep neural networks. For video modality, we use two kinds of visual features: the hand-crafted feature Improved Dense Trajectories (IDT) \cite{DBLP:conf/iccv/WangS13a} and the deeply-learned feature Convolutional 3D (C3D) visual features \cite{DBLP:conf/iccv/TranBFTP15}, to evaluate whether our method can generalize well to both the hand-crafted and deeply-learned video features.  For Improved Dense Trajectories (IDT) feature, we adopt the Locality-constrained Linear Coding (LLC) \cite{DBLP:conf/cvpr/WangYYLHG10} scheme to represent the IDTs by $5$ local bases, and the codebook size is set to be $2000$. To reduce the complexity when constructing the codebook, only $200$ IDTs are randomly selected from each video. This encoding scheme has been verified its effectiveness by \cite{DBLP:journals/ijcv/ZhuS14}. For C3D features, we extract the outputs of fc6 layer from the model pre-trained on the sports-1M dataset \cite{DBLP:conf/cvpr/KarpathyTSLSF14} and then averaged over the segments to form a $4096$-dimensional feature vectors. For semantic representations of the action class names, we use two widely used class attribute vectors: human labeled attributes \cite{DBLP:conf/cvpr/LampertNH09} and automatically learned distributed semantic representations word2vec \cite{DBLP:conf/nips/MikolovSCCD13}. We adopt the skip-gram model \cite{DBLP:conf/nips/MikolovSCCD13} trained on the Google News dataset ($¡Ö 100$ billion words) and represent each class name by an L2-normalized $300$ dimensional vector. For any multi-word class name (e.g. `walking with dog'), we generate its vector by accumulating the word vectors of each unique word \cite{DBLP:conf/emnlp/MilajevsKSP14}. For HMDB51 dataset, there is no publicly available attribute representations of the action classes. Hence, only word2vec is used for EAD$\rightarrow$HMDB51 dataset. For UCF101 dataset, $115$ dimensional attribute vectors are available. Therefore, both word2vec and class attribute are used on Stanford40$\rightarrow$UCF101, ASD$\rightarrow$UCF101 and BU101$\rightarrow$ UCF101 datasets.

Our DIVAFN method has six free parameters: $a$, $b$, $c$, $d$, $\beta$ and $\lambda$. The optimum values of them are determined by experiments, which will be discussed detailedly in Section \uppercase\expandafter{\romannumeral4}. D. Parameter $d$ is the vector length of domain-invariant representation from image, video keyframe  and video neural networks, which is set by using various values (e.g. $512$, $1024$, $2048$ and $4096$) to evaluate the performance. We fix the mini-batch size to be $64$, the iteration number to be $100$, and the learning rate to be $10^{-4}$. All experiments are repeated for five times and the average results are reported.

\subsection{Experimental Results}
We evaluate the performance of the DIVAFN on four datasets, shown in Tables \ref{Table 4}-\ref{Table 7}. In these tables, the SVM means directly using SVM to classify the target domain videos, following the methods \cite{DBLP:journals/ijcv/WangKSL13,DBLP:conf/iccv/WangS13a} for action recognition in videos. Therefore, we compare the DIVAFN with the SVM to evaluate whether our method can improve the video action recognition using knowledge from image modality. To evaluate whether directly adding the examples from the image modality into the video modality is effective, we directly use the concatenation of image and video features to train an SVM classifier to recognize the videos in target domain. Specifically, the image features are the outputs of the convolutional neural network CNN-F which has already been pre-trained on the ImageNet, the video features are the IDT or C3D features. And we consider this method as the baseline method. To evaluate whether the finetuned deep model from ImageNet could help video recognition or not, we also compare our DIVAFN with a simple transfer learning method which simply finetunes a model from an ImageNet pretrained network and use this network to classify video keyframes directly. Specifically, we utilize the image network with parameters initialized by ImageNet dataset and use the keyframes as the input, which has been verified its effectiveness by previous work \cite{DBLP:conf/nips/SimonyanZ14}. To make a fair comparison, we extract the features from the eighth layer of the image network and then use these features to train an SVM classifier for action recognition in videos. And we name this method finetuned. As shown in Table \ref{Table 4}-\ref{Table 7}, we observe the following.

The baseline method can achieve slightly better performance than the SVM on Stanford40$\rightarrow$UCF101, ASD$\rightarrow$UCF101 and EAD$\rightarrow$HMDB51 datasets. But on larger scale dataset BU101$\rightarrow$UCF101, the performance of the baseline method drops significantly and becomes even lower than that of the SVM. This shows that adding action images into action videos can improve recognition performance with limited extent only when the scale of the dataset is not so large. With the increasing number of the samples, the domain shift between image and video modalities will become significantly larger, which decreases the performance of the recognition. Therefore, effective domain adaptation and cross-modal feature fusion methods should be designed to address these problems.

\begin{table}[!t]\renewcommand\tabcolsep{1.5pt}
            \renewcommand{\arraystretch}{1.0}
             \caption{Average accuracies on Stanford40$\rightarrow$UCF101 dataset. The video feature is the IDT and the semantic feature is the word2vec.}
              \label{Table 4}
            \centering
            \begin{tabular}{|c||c|c|c||c|c|c|c|c|}\hline
           Ratio&SVM&baseline&finetuned&$d=512$&$d=1024$&$d=2048$&$d=4096$\\\hline
           $50\%$&$92.6$&$94.0$&$95.5$&$99.2$&$99.4$&$99.5$&$\mathbf{99.8}$\\\hline
           $40\%$&$92.1$&$93.3$&$94.1$&$98.4$&$98.9$&$99.2$&$\mathbf{99.6}$\\\hline
           $30\%$&$88.5$&$92.8$&$92.4$&$97.2$&$98.3$&$98.4$&$\mathbf{99.4}$\\\hline
           $20\%$&$87.1$&$89.4$&$89.1$&$97.2$&$98.0$&$98.1$&$\mathbf{98.1}$\\\hline
           $10\%$&$78.9$&$84.9$&$84.0$&$89.7$&$94.9$&$95.5$&$\mathbf{95.6}$\\\hline
            \end{tabular}
 \end{table}

  \begin{table}[!t]\renewcommand\tabcolsep{1.5pt}
            \renewcommand{\arraystretch}{1.0}
             \caption{Average accuracies on ASD$\rightarrow$UCF101 dataset. The video feature is the IDT and the semantic feature is the word2vec.}
              \label{Table 5}
            \centering
            \begin{tabular}{|c||c|c|c||c|c|c|c|c|}\hline
            Ratio&SVM&baseline&finetuned&$d=512$&$d=1024$&$d=2048$&$d=4096$\\\hline
           $50\%$&$85.0$&$87.6$&$81.2$&$94.0$&$94.4$&$95.2$&$\mathbf{97.2}$\\\hline
           $40\%$&$78.9$&$83.0$&$79.2$&$91.6$&$94.0$&$94.0$&$\mathbf{94.3}$\\\hline
           $30\%$&$76.1$&$80.0$&$75.8$&$88.0$&$88.5$&$90.5$&$\mathbf{92.8}$\\\hline
           $20\%$&$71.5$&$70.8$&$71.4$&$81.7$&$84.2$&$86.7$&$\mathbf{87.0}$\\\hline
           $10\%$&$60.8$&$65.1$&$61.3$&$76.0$&$76.8$&$77.5$&$\mathbf{78.2}$\\\hline
            \end{tabular}
 \end{table}

  \begin{table}[!t]\renewcommand\tabcolsep{1.5pt}
            \renewcommand{\arraystretch}{1.0}
             \caption{Average accuracies on EAD$\rightarrow$HMDB51 dataset. The video feature is the IDT and the semantic feature is the word2vec.}
              \label{Table 6}
            \centering
            \begin{tabular}{|c||c|c|c||c|c|c|c|c|}\hline
           Ratio&SVM&baseline&finetuned&$d=512$&$d=1024$&$d=2048$&$d=4096$\\\hline
           $50\%$&$61.6$&$70.1$&$60.6$&$71.7$&$72.6$&$74.6$&$\mathbf{76.2}$\\\hline
           $40\%$&$61.4$&$67.7$&$58.9$&$69.6$&$71.0$&$73.0$&$\mathbf{74.7}$\\\hline
           $30\%$&$56.0$&$64.1$&$56.1$&$68.0$&$68.6$&$70.0$&$\mathbf{71.1}$\\\hline
           $20\%$&$53.0$&$60.6$&$51.6$&$63.5$&$66.1$&$66.7$&$\mathbf{67.0}$\\\hline
           $10\%$&$42.8$&$50.7$&$44.4$&$52.5$&$56.3$&$56.6$&$\mathbf{60.1}$\\\hline
            \end{tabular}
 \end{table}

  \begin{table}[!t]\renewcommand\tabcolsep{1.5pt}
            \renewcommand{\arraystretch}{1.0}
             \caption{Average accuracies over three train/test splits for BU101$\rightarrow$UCF101 dataset. The video feature is the C3D and the semantic feature is the word2vec.}
              \label{Table 7}
            \centering
            \begin{tabular}{|c||c|c|c||c|c|c|c|c|}\hline
            Ratio&SVM&baseline&finetuned&$d=512$&$d=1024$&$d=2048$&$d=4096$\\\hline
           $100\%$&$81.7$&$79.1$&53.0&$84.6$&$86.7$&$87.4$&$\mathbf{88.4}$\\\hline
           $50\%$&$81.7$&$79.3$&51.4&$84.6$&$86.4$&$87.2$&$\mathbf{87.8}$\\\hline
           $33\%$&$81.0$&$78.2$&50.3&$84.1$&$85.3$&$86.2$&$\mathbf{86.6}$\\\hline
           $20\%$&$80.0$&$77.1$&47.4&$83.6$&$85.2$&$86.1$&$\mathbf{86.1}$\\\hline
           $10\%$&$76.5$&$73.5$&42.9&$80.0$&$81.4$&$82.1$&$\mathbf{82.1}$\\\hline
           $5\%$&$70.5$&$68.0$&35.8&$73.3$&$74.4$&$75.2$&$\mathbf{75.4}$\\\hline
            \end{tabular}
 \end{table}

\begin{figure}[!t]
\centering
\includegraphics[scale=0.315]{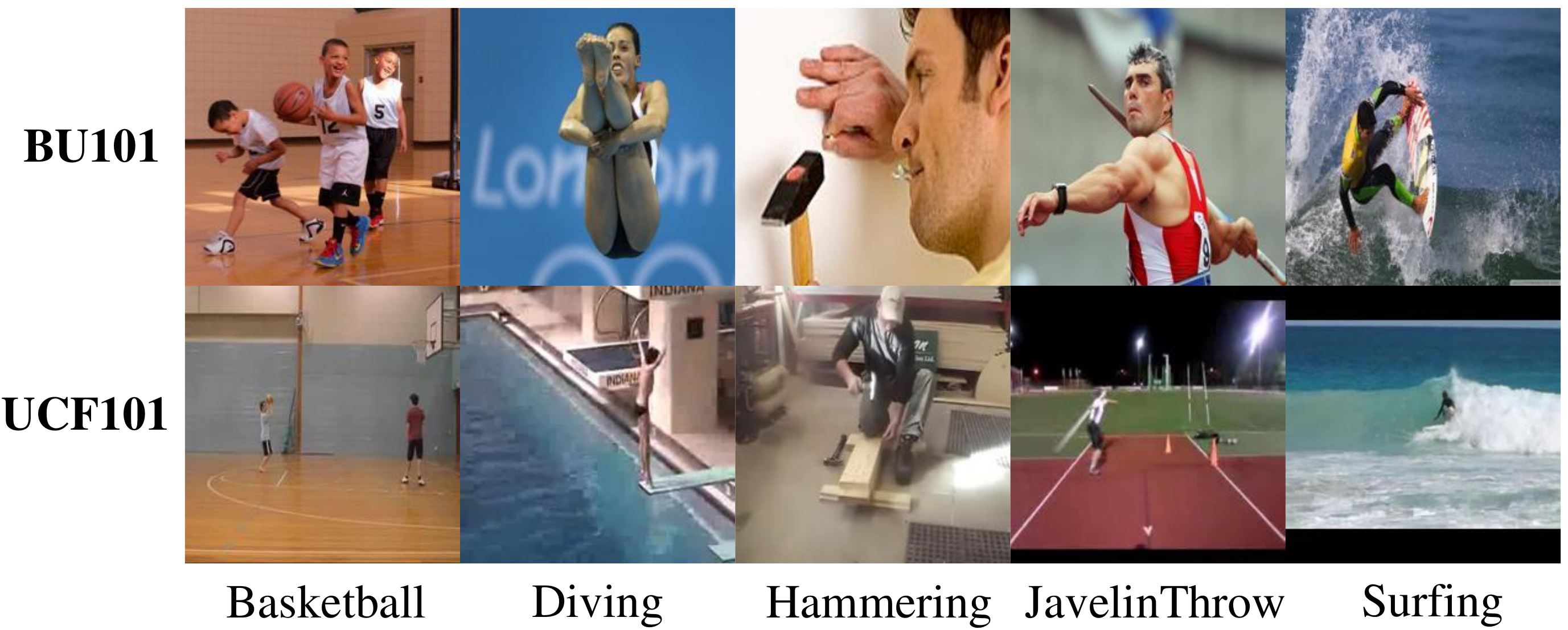}
\caption{Sample images from BU101 and UCF101 datasets. The first row shows images from BU101 image dataset. The second row shows images from UCF101 video dataset.}
\label{Fig4}
\end{figure}

The finetuned deep neural network only achieves comparative performance with the SVM on Stanford40$\rightarrow$UCF101 and ASD$\rightarrow$UCF101 datasets. But when addressing video action recognition problem on EAD$\rightarrow$HMDB51 and BU101$\rightarrow$UCF101 datasets, the performance drops significantly especially when the dataset is BU101$\rightarrow$UCF101. This is because that large modality difference between images and video exists on these large-scale action recognition datasets. To show the difference between images and videos, we show some example action images from the BU101 and the UCF101 datasets, which can be seen in Figure \ref{Fig4}. From Figure \ref{Fig4}, we can see that many of the action images are significantly different from video frames in camera viewpoint, lighting, human pose, and background. Action images tend to focus on representative moments and have more diverse examples of an action. In contrast, videos usually capture a whole temporal progression of an action and contain many redundant and uninformative frame. Therefore, simply using the finetuned model from image datasets to improve video action recognition performance without effective domain adaption and feature fusion methods is not effective.

From above methioned evaluation results of the baseline and the finetuned methods, we can see that the proposed DIVAFN outperforms the other methods in different settings. This proves that effective domain adaption and feature fusion methods can improve the performance of the image-to-video action recognition. It also validates that our DIVAFN can not noly effectively reduce the modality gap between images and videos but also make good use of the complemetary information between images and videos. Even when the number of training samples is limited, the improvement of the performance is still significant on four datasets. This advantage is especially desirable in real-world scenarios where annotated videos are scarce.

Because high dimensionality of the feature brings more informative and discriminative knowledges, the performance of the DIVAFN increases when the value of $d$ increases. This conclusion is valid when the feature's dimensionality stays in the reasonable scope. If the dimensionality exceeds the reasonable scope, the discriminability of the feature will degrade because of the redundant information. Since the domain-invariant representation is the output of the image, keyframe, and video networks, the last full-connected layer of the deep network decides the reasonable scope the feature's dimensionality. In our work, the length of the last full-connected layer of the deep image network and deep keyframe network is $4096$, and the deep video network is $8192$. As the dimensionality of the domain-invariant representation from images, keyframes and videos is set to be the same in our work, which should not exceed $4096$. Therefore, the reasonable scope of domain-invariant representation is $(0,4096]$. Based on the above analysis, we choose $d=4096$ as the dimension of the learned domain-invariant representations in the following experiments.

Although the BU101$\rightarrow$UCF101 dataset is a large-scale image-video dataset which contains significantly large domain shift between image and video modalities, our method can still significantly improve action recognition performance in videos across different number of training samples. This demonstrates that our method is effective on large-scale dataset.

\subsection{Parameter Sensitivity Analysis}

\begin{figure*}[!t] \centering
\subfloat[Stanford40$\rightarrow$UCF101] { \label{Fig5:a}
\includegraphics[width=0.485\columnwidth]{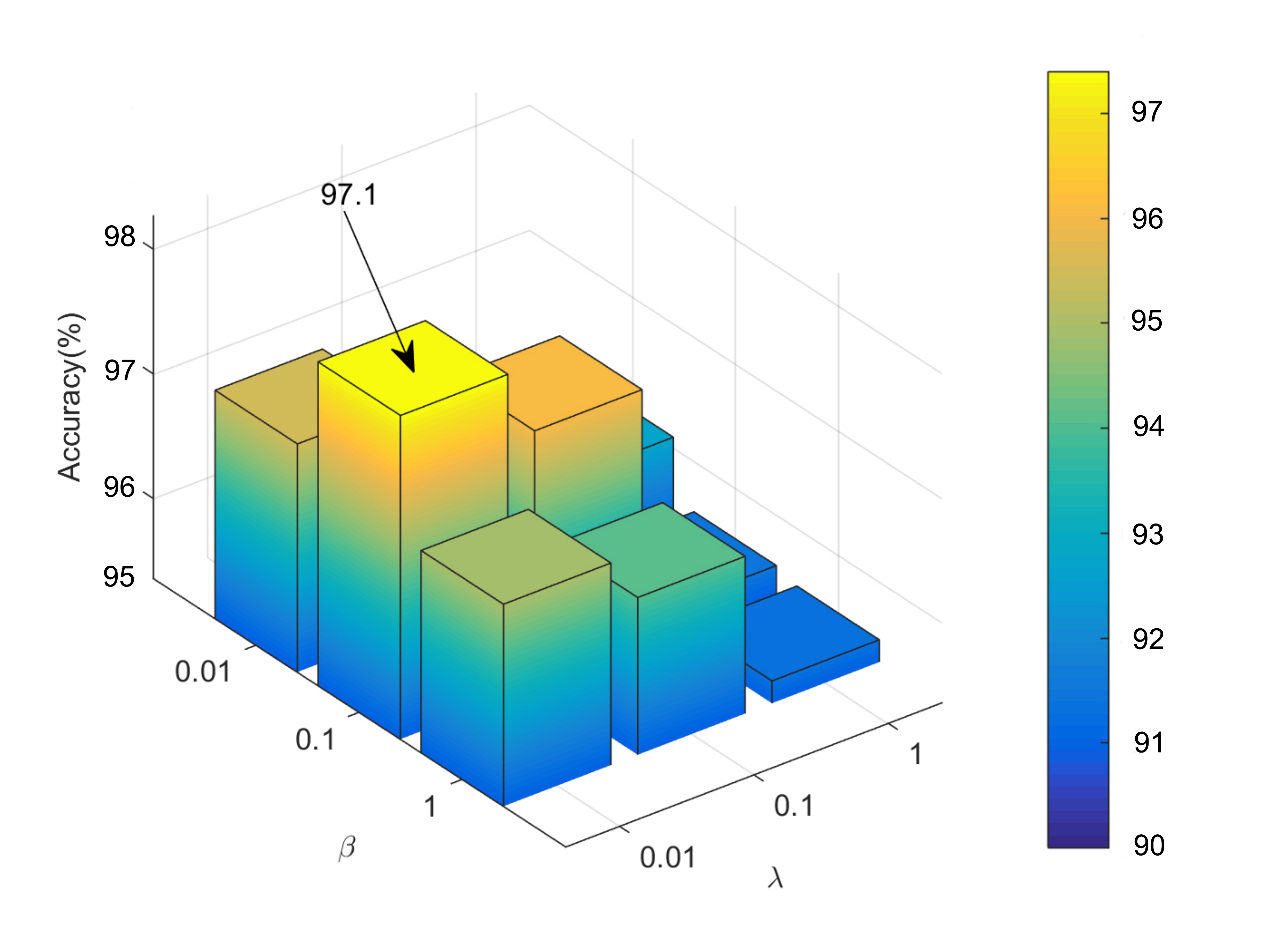}
}
\subfloat[ASD$\rightarrow$UCF101] { \label{Fig5:b}
\includegraphics[width=0.485\columnwidth]{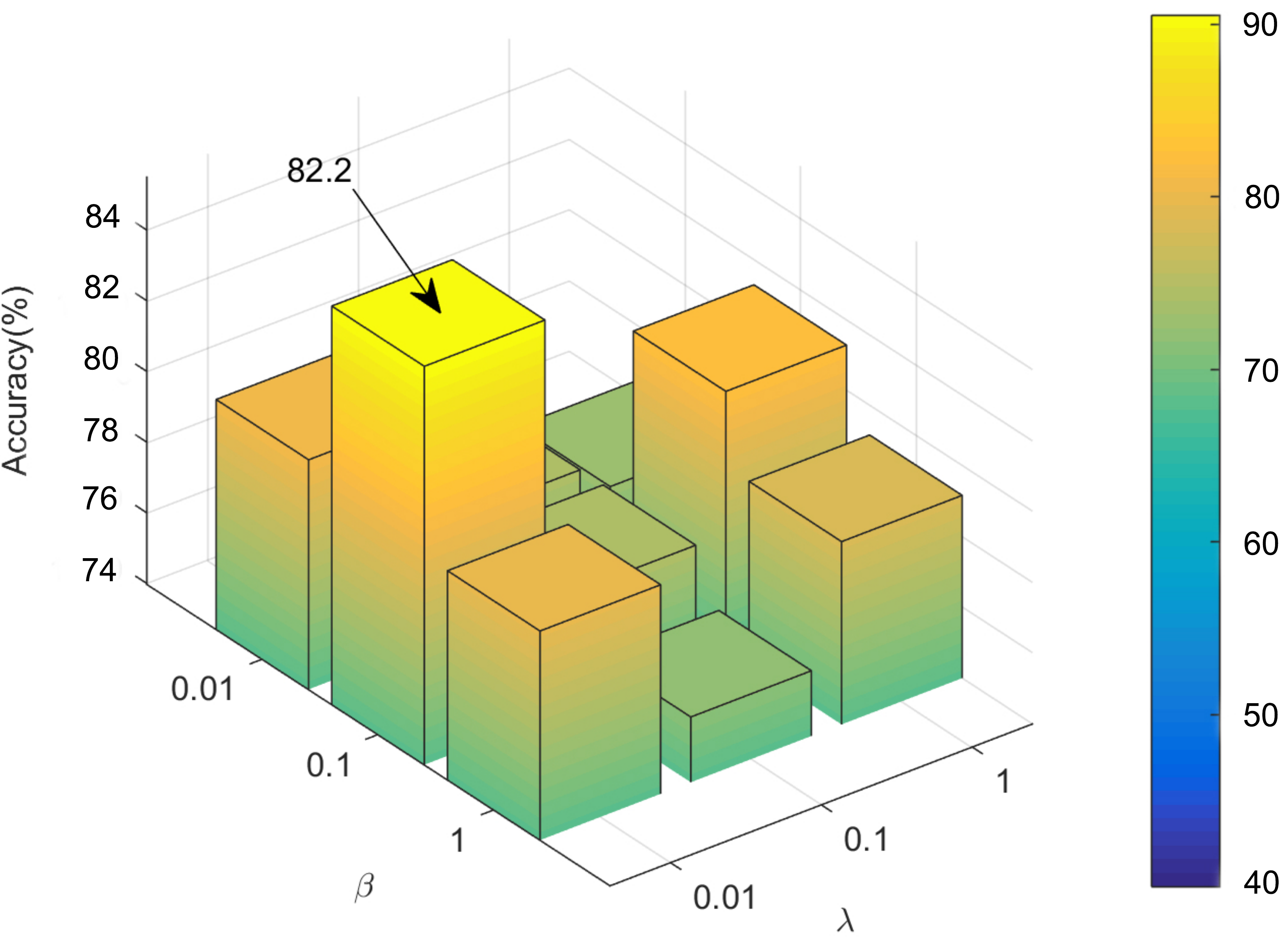}
}
\subfloat[EAD$\rightarrow$HMDB51] { \label{Fig5:c}
\includegraphics[width=0.485\columnwidth]{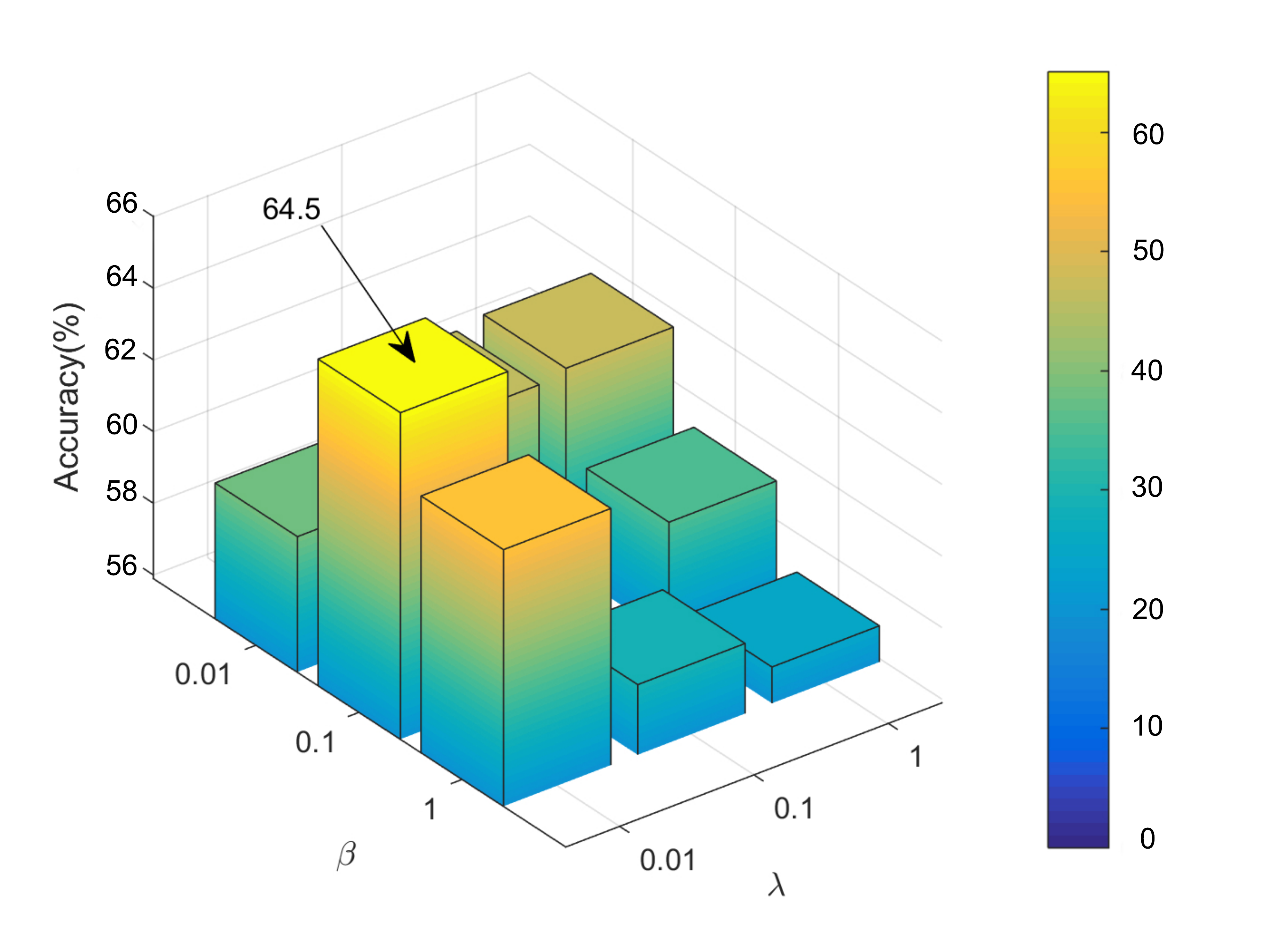}
}
\subfloat[BU101$\rightarrow$UCF101] { \label{Fig5:d}
\includegraphics[width=0.485\columnwidth]{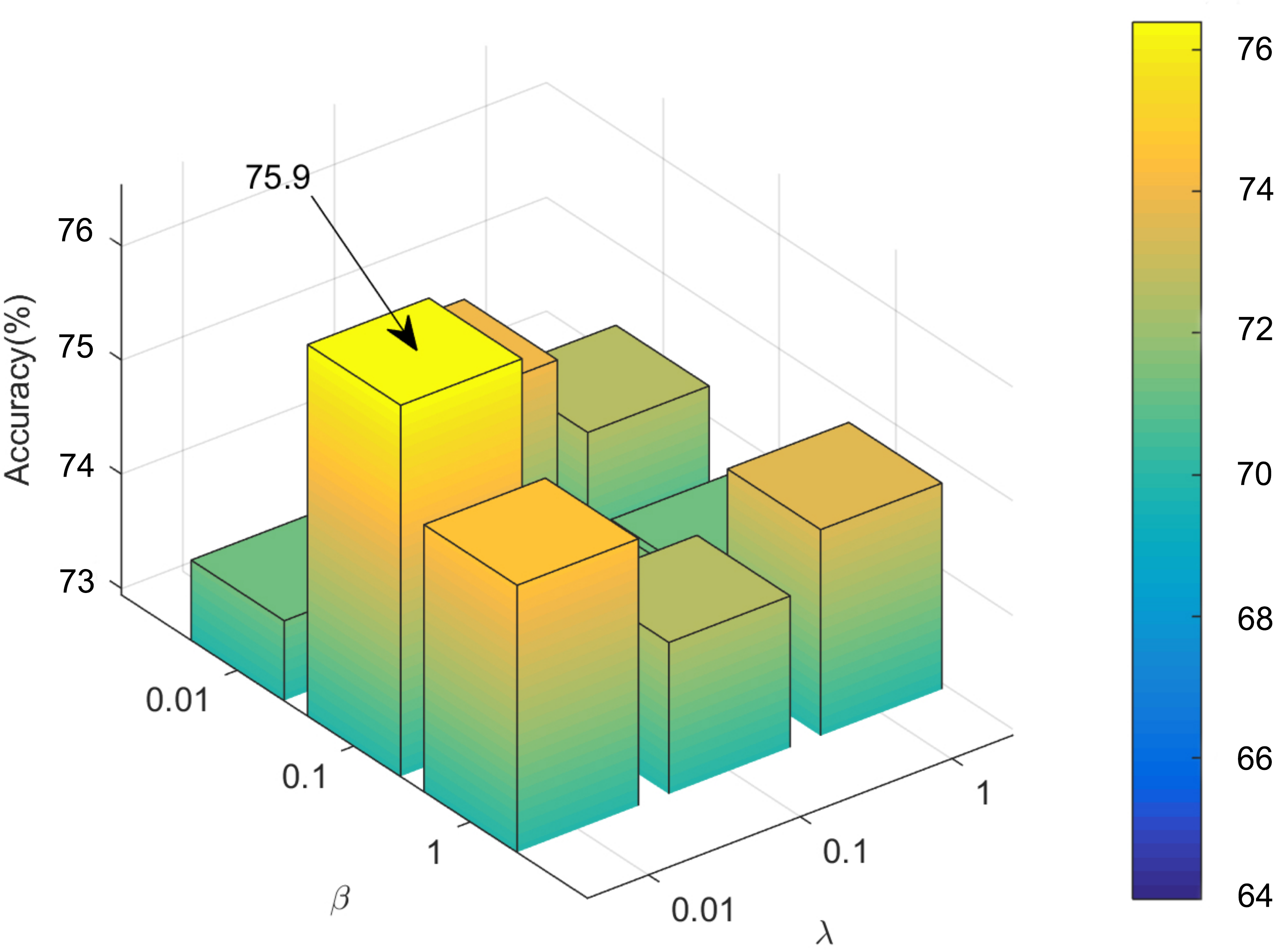}
}
\caption{ Recognition accuracies difference with respect to $\beta$ and $\lambda$ on four datasets. }
\label{Fig5}
\end{figure*}

There are five regularization parameters $a$, $b$, $c$, $\beta$ and $\lambda$ in Eq. (\ref{eq14}). To learn how they influence the performance, we conduct the parameter sensitivity analysis on four datasets when fixing $d=4096$. The video feature and semantic feature are C3D and word2vec, respectively. For Stanford40$\rightarrow$UCF101, ASD$\rightarrow$UCF101 and EAD$\rightarrow$HMDB51 datasets, we conduct experiments when the ratio of training samples is $10\%$. For BU101$\rightarrow$UCF101 dataset, we conduct experiments when the ratio of training samples is $5\%$ and use the train/test split$1$. The results of using other ratios of training samples and train/test splits are not given as the similar results.

\begin{figure*} [!t]\centering
\subfloat[Stanford40$\rightarrow$UCF101] { \label{Fig6:a}
\includegraphics[width=0.485\columnwidth]{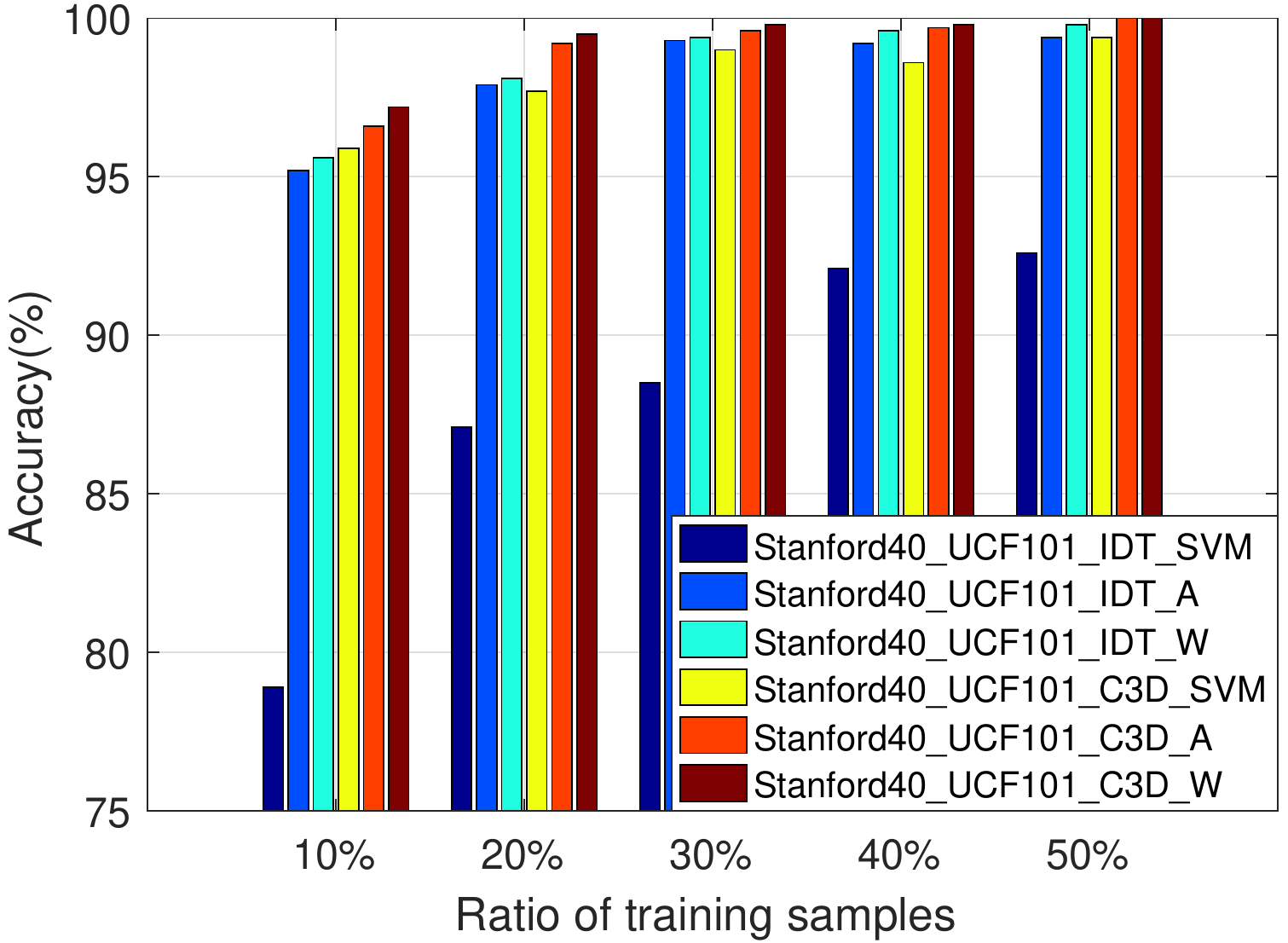}
}
\subfloat[ASD$\rightarrow$UCF101] { \label{Fig6:b}
\includegraphics[width=0.485\columnwidth]{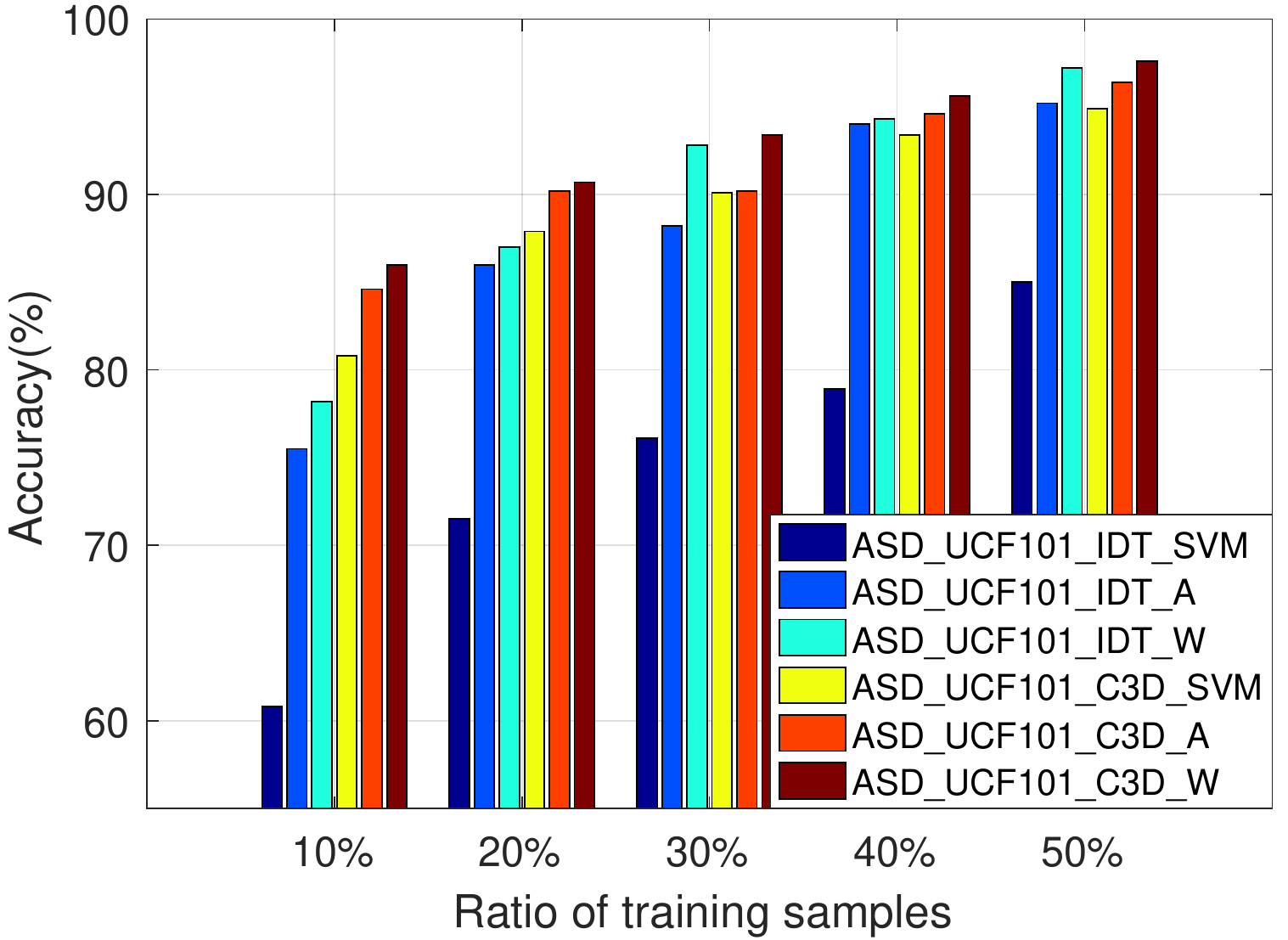}
}
\subfloat[EAD$\rightarrow$HMDB51] { \label{Fig6:c}
\includegraphics[width=0.485\columnwidth]{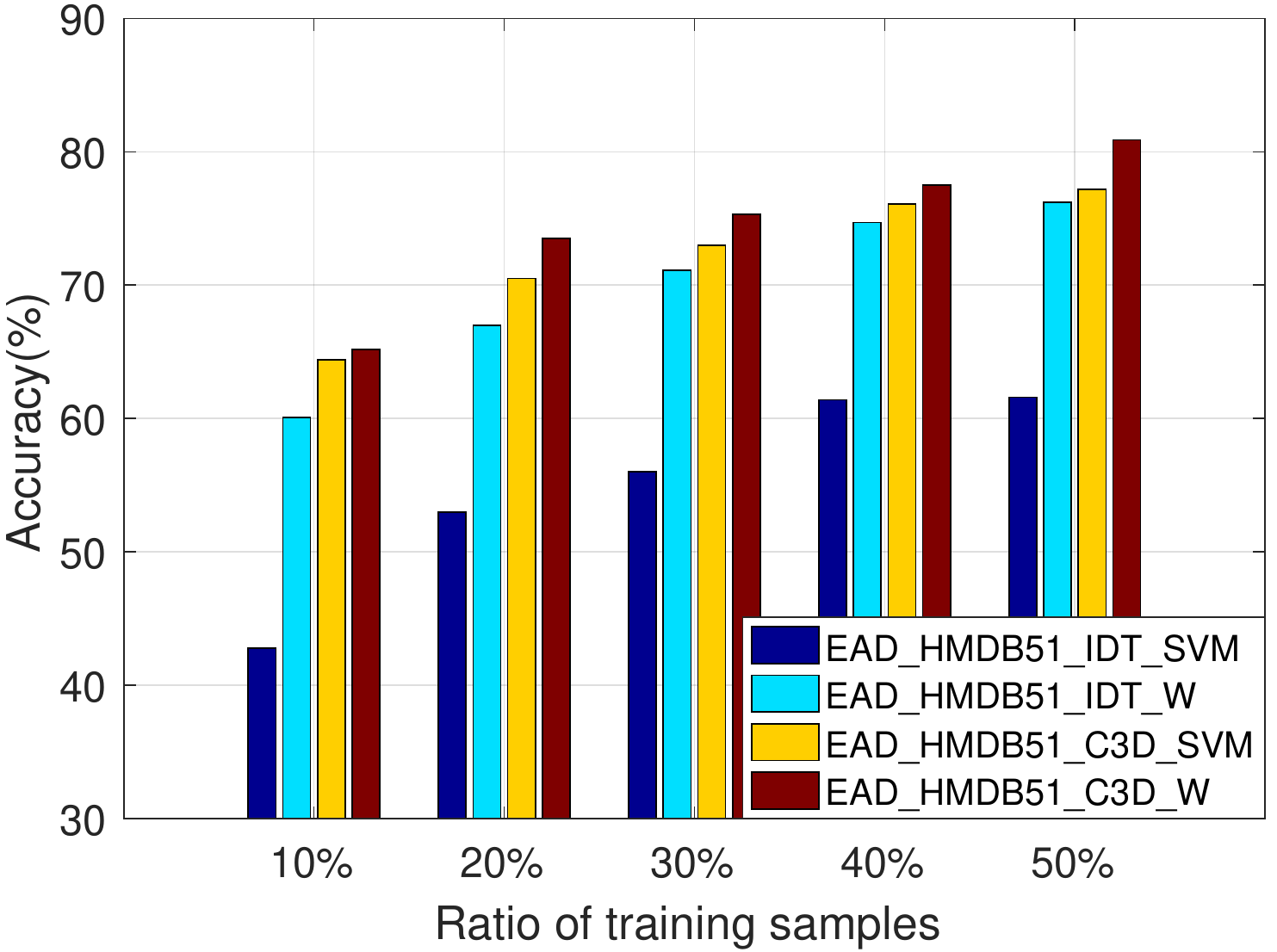}
}
\subfloat[BU101$\rightarrow$UCF101] { \label{Fig6:d}
\includegraphics[width=0.485\columnwidth]{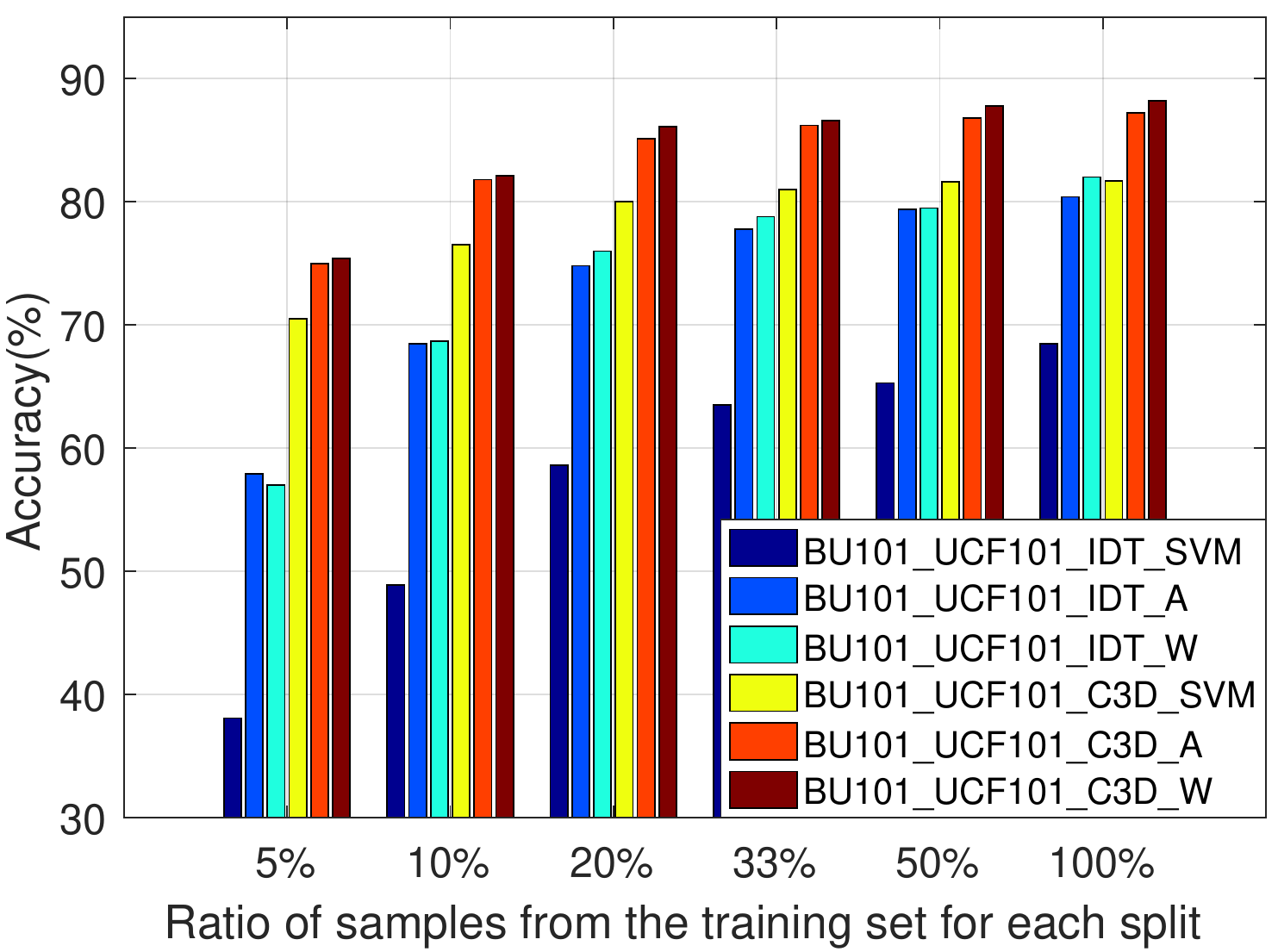}
}
\caption{ Comparison with the SVM on four datasets in terms of different video and semantic features. }
\label{Fig6}
\end{figure*}

Specifically, we first determine the optimum values of $\beta$ and $\lambda$ while setting $a=b=c=1$. From Fig. \ref{Fig5}, we can see that the best performance is achieved when $\beta=0.1, \lambda=0.01$ for all these four datasets. This means that the decoders should be attached more importance than the encoders. This is because the decoders of our autoencoders control the reconstruction process which makes the learned hidden feature representations be easily projected back to the original domain-invariant feature space. If we attach more importance to the decoders, the hidden unit can be encouraged to be a good representation of the inputs as the reconstruction process captures the intrinsic structure of the input domain-invariant features. Nevertheless, the importance of the encoders should not be neglected since they are also important to encourage the features in a semantic space to contain both learned domain-invariant information and semantic knowledge by jointly learning encoders and decoders. Therefore, we choose $\beta=0.1, \lambda=0.01$ in the following experiments.

When considering an extreme condition that setting the values of $\beta$ and $\lambda$ to zero, our method becomes domain-invariant representations learning part of the proposed DIVAFN according to Eq. (\ref{eq14}). This is also equivalent to ignoring the cross-modal feature fusion part of our DIVAFN. In the following Ablation Study section (Section \uppercase\expandafter{\romannumeral4}.F), we will evaluate the performance of our method when ignoring the cross-modal feature fusion part, named DIVA.

\begin{table}[!t]
            \renewcommand{\arraystretch}{1.0}
             \caption{The relationship between parameters $a$, $b$, $c$ and the action recognition accuracy (\%) on four datasets. The three numbers in the bracket are the values of $a$, $b$ and $c$.}
              \label{Table 8}
            \centering
            \begin{tabular}{|c|c|c|c|c|c|c|c|c|c|}\hline
          $(a,b,c)$&$\beta$&$\lambda$&S$\rightarrow$U&A$\rightarrow$U&E$\rightarrow$H&B$\rightarrow$U\\\hline
           $(1,1,1)$&$0.1$&$0.01$&$96.4$&$78.6$&$62.2$&$75.6$\\\hline
           $(1,1,0.1)$&$0.1$&$0.01$&$95.5$&$78.6$&$62.5$&$75.7$\\\hline
           $(1,0.1,1)$&$0.1$&$0.01$&$ 96.4$&$78.2$&$63.1$&$74.9$\\\hline
           $(1,0.1,0.1)$&$0.1$&$0.01$&$96.4$&$79.1$&$63.4$&$76.0$\\\hline
           $(0.1,1,1)$&$0.1$&$0.01$&$96.3$&$77.5$&$63.5$&$75.9$\\\hline
           $(0.1,1,0.1)$&$0.1$&$0.01$&$96.4$&$77.7$&$61.9$&$75.2$\\\hline
           $(0.1,0.1,1)$&$0.1$&$0.01$&$\mathbf{96.5}$&$\mathbf{79.7}$&$\mathbf{64.0}$&$\mathbf{76.3}$\\\hline
           $(0.1,0.1,0.1)$&$0.1$&$0.01$&$96.4$&$78.6$&$62.5$&$75.0$\\\hline
            \end{tabular}
 \end{table}

Then, we determine the optimum values of parameters $a$, $b$ and $c$ while setting $\beta=0.1, \lambda=0.01, d=4096$. Table \ref{Table 8} shows the performance of our DIVAFN on four datasets by allocating different values to the parameters $a$, $b$ and $c$. Actually, $a$, $b$ and $c$ denote the weighting coefficients controlling the importance of the three similarity related functions: the similarity between image and video modalities, the similarity between image and keyframe modalities, and the similarity between keyframe and video modalities. From Table \ref{Table 8}, we can see that the best accuracies are achieved when $a=0.1, b=0.1, c=1$ for all four datasets, which shows the following. 1) The optimum values of $a$, $b$  are equal to that of $\beta$, which means that both domain-invariant feature learning and semantic learning should be attached the same importance. In this way, we can obtain more discriminative and robust representations which contain both learned domain-invariant information and semantic knowledge. 2) The optimum values of $c$ is larger than that of $a$, $b$ and $\beta$, this means that we should attach more importance to the similarity between keyframe and video modalities than other two similarities and the semantic learning. This is because keyframes are essentially images, and the modality gap between keyframes and images is smaller than that between keyframes and videos. Moreover, the keyframe modality is the bridge to utilize the relationship between image and video modalities. If we pay more importance on reducing the modality shift between keyframe and video, we can obtain good domain-invariant representations in all modalities including images, keyframes and videos.

\subsection{Influence of Different Video and Semantic Features}

In this experiment, we evaluate the performance difference of our proposed DIVAFN by using different types of video features and semantic features. The video features include the hand-crafted feature the Improved Dense Trajectories (IDT) \cite{DBLP:conf/iccv/WangS13a} and the deeply-learned feature Convolutional 3D (C3D) visual features \cite{DBLP:conf/iccv/TranBFTP15}. The semantic features of action class names include human labeled attributes \cite{DBLP:conf/cvpr/LampertNH09} and automatically learned distributed semantic representations word2vec \cite{DBLP:conf/nips/MikolovSCCD13}. Therefore, we get four different feature combinations, i.e., IDT\_A, IDT\_W, C3D\_A, and C3D\_W. Here A represents the human annotated class attribute vectors and W represents the word2vec embedding. Since there are two kinds of video features IDT and C3D, we have two baseline methods named IDT\_SVM and C3D\_SVM. To make an intuitive comparison with the SVM on four datasets using four different feature combinations, we show the results in Fig. \ref{Fig6}.

Generally, the performance of using C3D as the video feature is better than that of using IDT because the C3D is the deeply learned feature which may be more discriminative than hand-crafted feature IDT. The performance of using word2vec as semantic feature is slightly better than that of using human labeled attributes when the video features are the same. This is because that the higher dimension of word2vec features make the learned hidden representations contain more informative and discriminative knowledge. In addition, the marginal performance difference between annotated attribute and word2vec indicates that they are both effective in our proposed DIVAFN. All four feature combinations achieve significantly better performance than that of the SVM on four datasets, which validates the effectiveness of the video features and semantic features. Most importantly, our DIVAFN can effectively enhance recognition performance in videos by transferring knowledge from image modality.

\subsection{Ablation Study}
To evaluate the importance of the domain-invariant representations learning part and cross-modal feature fusion part, we construct three new algorithms based on the proposed DIVAFN method.

Keyframe-Video Concatenation (KVC), which is the simply concatenation of raw keyframe feature and video feature. Since the input of keyframe modality is the image pixels, KVC uses the output from the eighth layer of the keyframe network which has not been trained by the cross-modal similarities metric as the keyframe feature. And the video feature is IDT or C3D. Then, the simply concatenation of keyframe and video features is used to trained an SVM classifier for action recognition in videos.

Deep Image-to-Video Adaptation (DIVA), which is the domain-invariant representations learning part of the DIVAFN. DIVA uses the output from the eighth layer of the keyframe network which has been trained by the cross-modal similarities metric as the keyframe feature. DIVA uses the output from the second layer of the video network which has been trained by the cross-modal similarities metric as the video feature. Then, DIVA uses the simply concatenations of the keyframe feature and video feature as the final fused feature for action recognition in videos by the SVM classifier.

Deep Image-to-Video Fusion (DIVF), which is the cross-modal feature fusion part of the DIVAFN. DIVF uses the output from the eighth layer of the keyframe network which has not been trained by the cross-modal similarities metric as the keyframe feature. Because the video network is only randomly initialized, which is not suitable for video feature extraction. Thus, DIVF uses IDT or C3D as the video feature. Then, the keyframe feature, the video feature, and their concatenations are projected into the same semantic space by learning three autoencoders Auto-H, Auto-G and Auto-E, respectively. Finally, the concatenation of the learned semantic feature representations from these three autoencoders is used to train an SVM classifier for action recognition in videos.

We conduct experiments on Stanford40$\rightarrow$UCF101 and ASD$\rightarrow$UCF101 datasets. For Stanford40$\rightarrow$UCF101 dataset, we use IDT and word2vec as the video feature and semantic feature, respectively. For ASD$\rightarrow$UCF101 dataset, we use IDT and attribute. The results are shown in Table \ref{Table 9} and Table \ref{Table 10}. Compared with the SVM, KVC achieves slightly better performance on two datasets. This demonstrates that the simply combination of image and video features can improve action recognition performance in videos due to the complementarity. However, the performance improvement is not significant due to the existence of domain shift between image modality and video modality. Both DIVA and DIVF perform better than KVC and SVM, this validates the effectiveness of our domain-invariant learning method and cross-modal feature fusion method. Among various methods and ratios of training samples, DIVAFN achieves the best performance on two datasets. This verifies that we can obtain more compact, informative and discriminative feature representations to enhance action recognition in videos by integrating domain-invariant representations learning and cross-modal feature fusion into a unified deep learning framework.

\begin{table}[!t]
            \renewcommand{\arraystretch}{1.0}
             \caption{Average accuracies on Stanford40$\rightarrow$UCF101 dataset using the SVM, KVC, DIVA, DIVF and DIVAFN methods.}
              \label{Table 9}
            \centering
            \begin{tabular}{|c|c|c|c|c|c|c|}\hline
           Ratio&SVM&KVC&DIVA&DIVF&DIVAFN\\\hline
           $50\%$&$92.6$&$96.7$&$98.1$&$98.9$&$\mathbf{99.8}$\\\hline
           $40\%$&$92.1$&$96.3$&$98.4$&$99.0$&$\mathbf{99.6}$\\\hline
           $30\%$&$88.5$&$94.9$&$97.2$&$98.7$&$\mathbf{99.4}$\\\hline
           $20\%$&$87.1$&$93.4$&$96.1$&$97.9$&$\mathbf{98.1}$\\\hline
           $10\%$&$78.9$&$86.8$&$91.8$&$95.4$&$\mathbf{95.6}$\\\hline
            \end{tabular}
 \end{table}

  \begin{table}[!t]
            \renewcommand{\arraystretch}{1.0}
             \caption{Average accuracies on ASD$\rightarrow$UCF101 dataset using the SVM, KVC, DIVA, DIVF and DIVAFN methods.}
              \label{Table 10}
            \centering
            \begin{tabular}{|c|c|c|c|c|c|c|}\hline
            Ratio&SVM&KVC&DIVA&DIVF&DIVAFN\\\hline
           $50\%$&$85.0$&$84.2$&$91.2$&$90.0$&$\mathbf{95.2}$\\\hline
           $40\%$&$78.9$&$82.5$&$92.0$&$89.0$&$\mathbf{94.0}$\\\hline
           $30\%$&$76.1$&$78.2$&$89.7$&$82.1$&$\mathbf{88.2}$\\\hline
          $20\%$&$71.5$&$73.2$&$80.7$&$77.9$&$\mathbf{86.0}$\\\hline
          $10\%$&$60.8$&$61.2$&$72.6$&$67.8$&$\mathbf{75.5}$\\\hline
                      \end{tabular}
 \end{table}

\subsection{Comparison with Domain Adaptation Methods}
\renewcommand\tabcolsep{2.0pt}\renewcommand\tabcolsep{2.5pt}
\begin{table*}[!t]
            \renewcommand{\arraystretch}{1}
             \caption{Comparison with other domain adaptation methods. The five accuracy numbers in the bracket are the average recognition accuracies with the ratios of training samples: $10\%$, $20\%$, $30\%$, $40\%$ and $50\%$, respectively. The six accuracy numbers in the bracket are the average recognition accuracies over three train/test splits with the ratios of training samples for each split: $5\%$, $10\%$, $20\%$, $33\%$, $50\%$ and $100\%$, respectively. }
             \label{Table 11}
            \centering
            \begin{tabular}{|c|c|c|c|c|}\hline
            Method&Stanford40$\rightarrow$UCF101&ASD$\rightarrow$UCF101&EAD$\rightarrow$HMDB51&BU101$\rightarrow$UCF101\\\hline
           SVM&$(95.9,97.7,99.0,98.6,99.4)$&$(80.8,87.9,90.1,93.4,94.9)$&$(64.4,70.5,73.0,76.1,77.2)$&$(70.5,76.5,80.0,81.0,81.7,81.7)$\\
           TCA \cite{DBLP:journals/tnn/PanTKY11}&$(95.2,97.3,98.7,99.2,99.1)$&$(77.7,84.1,88.9,91.1,93.0)$&$(62.5,67.1,69.5,72.1,74.4)$&$(73.0,78.9,81.4,81.8,82.2,82.8)$ \\
           GFK \cite{DBLP:conf/cvpr/GongSSG12}&$(88.0,93.3,96.4,96.7,98.5)$&$(65.6,71.6,76.9,81.3,83.6)$&$(51.0,58.4,62.3,63.8,66.7)$&$(67.0,71.9,73.5,74.4,75.6,76.3)$\\
           JDA \cite{DBLP:conf/iccv/LongWDSY13}&$(97.0,98.4,98.5,99.2,99.0)$&$(77.6,84.4,89.6,90.0,92.1)$&$(60.7,68.9,71.2,72.3,73.9)$&$(73.0,78.9,81.4,81.8,82.2,82.8)$ \\
           TJM \cite{DBLP:conf/cvpr/LongWDSY14}&$(95.7,96.9,98.3,98.6,98.9)$&$(79.2,85.0,88.5,90.0,93.5)$&$(\mathbf{65.6},68.5,72.6,74.8,75.1)$&$(37.9,46.6,53.4,57.4,59.0,61.1)$\\
           CORAL \cite{DBLP:conf/aaai/SunFS16}&$(91.7,97.0,98.1,98.6,99.4)$&$(66.2,71.6,83.8,86.2,90.0)$&$(58.4,66.5,69.9,72.5,75.6)$&$(74.4,81.4,83.8,84.8,85.2,85.5)$\\
           JGSA \cite{DBLP:conf/cvpr/ZhangLO17}&$(\mathbf{98.7},99.2,99.6,99.7,99.7)$&$(80.4,89.0,91.8,95.0,96.6)$&$(62.9,71.4,73.7,75.6,76.5)$&$(68.1,72.6,75.9,77.2,77.6,77.6)$ \\
           MEDA \cite{DBLP:conf/mm/WangFCYHY18}&$(94.1,95.7,94.8,97.1,97.2)$&$(68.8,70.2,79.1,79.0,82.8)$&$(54.5,61.5,62.9,65.8,68.8)$&$(66.8,72.7,75.9,76.6,77.3,77.9)$\\\hline
          DIVAFN&$(97.2,\mathbf{99.5},\mathbf{99.8},\mathbf{99.8},\mathbf{100})$&$(\mathbf{86.0},\mathbf{90.7},\mathbf{93.4},\mathbf{95.6},\mathbf{97.6})$&$(65.2,\mathbf{73.5},\mathbf{75.3},\mathbf{77.5},\mathbf{80.9})$&$(\mathbf{75.4},\mathbf{82.1},\mathbf{86.1},\mathbf{86.6},\mathbf{87.8},\mathbf{88.4})$\\\hline
            \end{tabular}
 \end{table*}

 In this section, we evaluate whether our generated features for keyframe and video modalities are more transferable than other domain adaptation methods. An introduction of the comparison domain adaptation methods is as follows.
 \subsubsection{SVM} The most widely used and robust supervised recognition method for video action recognition, which directly classify target domain videos based on video action features.
 \subsubsection{TCA \cite{DBLP:journals/tnn/PanTKY11}}A domain adaptation method built with a dimensionality reduction framework to reduce the distance between domains in a latent space.
 \subsubsection{GFK \cite{DBLP:conf/cvpr/GongSSG12}}An unsupervised domain adaptation method that models domain shift by integrating an infinite number of subspaces that characterize changes in geometric and statistical properties from the source to the target domain.
 \subsubsection{JDA \cite{DBLP:conf/iccv/LongWDSY13}}A domain adaptation method that simultaneously reduces the difference in both the marginal and conditional distributions between source and target domains.
 \subsubsection{TJM \cite{DBLP:conf/cvpr/LongWDSY14}}An unsupervised domain adaptation method that reduces the domain difference by jointly matching the features and reweighting the
instances across domains in a principled dimensionality reduction procedure.
 \subsubsection{CORAL \cite{DBLP:conf/aaai/SunFS16}}An unsupervised domain adaptation method that minimizes the domain shift by aligning the second-order statistics of source and target distributions.
 \subsubsection{JGSA \cite{DBLP:conf/cvpr/ZhangLO17}}An unsupervised domain adaptation method that reduces the shift between domains by learning two coupled projections that project the source domain and target domain data into lowdimensional subspaces where the geometrical shift and distribution shift are reduced simultaneously.
 \subsubsection{MEDA \cite{DBLP:conf/mm/WangFCYHY18}}A domain adaptation method that learns a domain-invariant classifier in Grassmann manifold with structural risk minimization, while performing dynamic distribution alignment to quantitatively account for the relative importance of marginal and conditional distributions.
  \subsubsection{DIVAFN} Our proposed image-to-video adaptation and fusion method that integrates domain-invariant representations learning and cross-modal feature fusion into a unified deep learning framework.

 For these comparison domain adapatation methods, we use the codes shared by authors, and the values of parameters in these methods are selected based on the default setting. The source domain
 and target domain for these domain adaptation methods are images and videos, respectively. Specifically, the output from the eighth layer of the image network which has not been trained by the cross-modal similarities metric is used as the input of the source domain, and the video feature is used as the input
 of the target domain. Since we focus on the effectiveness of these domain adaptation methods, we only extract the features generated by these comparison methods. After domain adaptation, we use the concatenation of the generated image and video features to train an SVM  classifier to recognize the actions in videos. In order to have a fair comparison, we use the same experimental setting as our DIVAFN and report the best results.

 As shown in Table \ref{Table 11}, most of the domain adaptation methods performs worse than the SVM due to the modality shift between image and video modalities without effective domain adaptation and feature fusion strategies. Although the JGSA \cite{DBLP:conf/cvpr/ZhangLO17} can achieve better performance than the SVM on the first three datasets, its performance drops significantly when the dataset is BU101$\rightarrow$UCF101. This is because the BU101$\rightarrow$UCF101 dataset is a more challenge dataset which contains more action classes and samples. And thus the domain difference between image and video modalities is significant. Nevertheless, our DIVAFN performs significantly better on four datasets across different ratios of training samples. This shows that our DIVAFN can effectively reduce domain shift between image and video modalities on different datasets by our domain-invariant representations learning method. In addition, the action recognition performance in videos can be significantly improved by our newly designed cross-modal feature fusion method. Owing to the strengths of domain-invariant feature learning and cross-modal feature fusion, our DIVAFN achieves the best performance.

\subsection{Comparison with Action Recognition Methods}

  \begin{table}[!t]\renewcommand\tabcolsep{5pt}
            \renewcommand{\arraystretch}{1}
             \caption{Comparison with state-of-the-art methods on UCF101 dataset. The accuracy is averaged over three splits.}
              \label{Table 12}
            \centering
            \begin{tabular}{|c|c|}\hline
            Method&Accuracy (\%)\\\hline
            IDT+LLC encoding+linear SVM (baseline)&$68.5$\\
            C3D+linear SVM (baseline)&$81.7$\\\hline
            Dynamic Image Nets \cite{DBLP:conf/cvpr/BilenFGVG16} &$76.9$\\
           C3D (1 net) + linear SVM \cite{DBLP:conf/iccv/TranBFTP15}&$82.3$\\
           Long-term recurrent ConvNet\cite{DBLP:journals/pami/DonahueHRVGSD17}&$82.9$\\
           Composite LSTM Model \cite{DBLP:conf/icml/SrivastavaMS15}&$84.3$\\
           C3D (3 nets) + linear SVM \cite{DBLP:conf/iccv/TranBFTP15}&$85.2$\\
           IDT-FV \cite{DBLP:conf/iccv/WangS13a}&$87.9$\\
           Two-stream CNN \cite{DBLP:conf/nips/SimonyanZ14}&$88.0$\\
          $F_{ST}$CN \cite{DBLP:conf/iccv/SunJYS15}&$88.1$\\
           LTC \cite{DBLP:journals/pami/VarolLS18}&$91.7$\\
           KVMF \cite{DBLP:conf/cvpr/ZhuHSCQ16}&$93.1$\\
           TSN(3 modalities) \cite{DBLP:conf/eccv/WangXW0LTG16}&$93.4$\\
           ATW\cite{DBLP:conf/ifip12/ZangWLZHZ18}&$94.6$\\
           Two-Stream I3D, Imagenet+Kinetics pre-training \cite{DBLP:conf/cvpr/CarreiraZ17}&$\mathbf{98.0}$\\\hline
           DIVAFN ($20\%$ training samples)&$86.1$\\
           DIVAFN ($33\%$ training samples)&$86.6$\\
           DIVAFN ($50\%$ training samples)&$87.8$\\
           DIVAFN ($100\%$ training samples)&$88.4$\\\hline
            \end{tabular}
 \end{table}

In this section, we compare our DIVAFN with other state-of-the-art video action recognition methods on UCF101 dataset. For DIVAFN, we use C3D and word2vec as the video feature and semantic feature, respectively. And we report the average accuracies across three train/test splits in Table \ref{Table 12}. With only $20\%$ of training videos, our DIVAFN achieves comparable or even better performance than some of the comparison methods. This shows the effectiveness of our DIVAFN when the training videos are limited. Using all training videos, our method achieves the accuracy
of $88.4$, which is comparable or even better than some of the deep learning based methods. To be noticed, some recent deep learning methods LTC \cite{DBLP:journals/pami/VarolLS18}, KVMF \cite{DBLP:conf/cvpr/ZhuHSCQ16},
TSN \cite{DBLP:conf/eccv/WangXW0LTG16}, ATW\cite{DBLP:conf/ifip12/ZangWLZHZ18}  and Two-Stream I3D \cite{DBLP:conf/cvpr/CarreiraZ17} can even achieve the  performance higher than $90\%$ on UCF101 dataset. However,
these deep learning methods usually require significantly large amount of training video samples to learn a robust classification model. Different from these promising deep learning methods, our method DIVAFN can improve video action recognition
performance by transferring knowledge from images to videos and conconcurrently reduce the amount of required training video samples.

To show the computational efficiency of our DIVAFN, we conduct experiments on four datasets with different ratios of training samples, and use C3D and word2vec as video feature and semantic feature, respectively. On Intel (R) CoreTM i7 system with 32GB RAM, NVIDIA GTX1070Ti GPU, and un-optimized Matlab code excluding the process of IDT or C3D video feature extraction, we evaluate average run-time for each video. For Stanford40$\rightarrow$UCF01, ASD$\rightarrow$UCF101 and EAD$\rightarrow$HMDB51 datasets, the average computational time with the ratios of training samples: $10\%$, $20\%$, $30\%$, $40\%$ and $50\%$ are: ($0.71$, $0.80$, $0.93$, $1.04$, $1.21$) seconds, ($1.41$, $1.49$, $1.57$, $1.74$, $1.78$) seconds and ($0.57$, $0.62$, $0.89$, $1.03$, $1.20$) seconds, respectively. For BU101$\rightarrow$UCF101 dataset, the average computational time over three train/test splits with the ratios of training samples for each split: $5\%$, $10\%$, $20\%$, $33\%$, $50\%$ and $100\%$ are ($0.11$, $0.21$, $0.36$, $0.67$, $1.07$, $2.49$) seconds.

\section{Conclusion}
In this paper, we propose a novel method, named Deep Image-to-Video Adaptation and Fusion Networks (DIVAFN), to enhance video action recognition by transferring knowledge from images. Our approach integrates domain-invariant representations learning and cross-modal feature fusion into a unified deep learning framework. To perform domain-invariant feature representations learning, we design three deep neural networks, one for image modality, one for keyframe modality, and the other for video modality, and design a novel cross-modal similarities metric to reduce the modality shift among them. To utilize the semantic relationship among images, keyframes and videos, we adopt an autoencoder architecture to obtain more compact, informative and discriminative representations from the hidden layer of the autoencoder. To effectively fuse keyframe and video feature representations, we simultaneously project learned domain-invariant keyframe features, video features and their concatenations to the same semantic space by learning three semantic autoencoders. And the concatenation of the learned semantic representations from these three autoencoders is used to train the classifier for action recognition in videos. Experimental results on four real-world datasets show the effectiveness of our DIVAFN in  improving action recognition performance in videos by transferring knowledge from images including the case when the number of training videos is limited.
\bibliographystyle{IEEEtran}
\bibliography{IEEEabrv,bibfile}

\begin{thebibliography}{10}
\providecommand{\url}[1]{#1}
\csname url@samestyle\endcsname
\providecommand{\newblock}{\relax}
\providecommand{\bibinfo}[2]{#2}
\providecommand{\BIBentrySTDinterwordspacing}{\spaceskip=0pt\relax}
\providecommand{\BIBentryALTinterwordstretchfactor}{4}
\providecommand{\BIBentryALTinterwordspacing}{\spaceskip=\fontdimen2\font plus
\BIBentryALTinterwordstretchfactor\fontdimen3\font minus
  \fontdimen4\font\relax}
\providecommand{\BIBforeignlanguage}[2]{{%
\expandafter\ifx\csname l@#1\endcsname\relax
\typeout{** WARNING: IEEEtran.bst: No hyphenation pattern has been}%
\typeout{** loaded for the language `#1'. Using the pattern for}%
\typeout{** the default language instead.}%
\else
\language=\csname l@#1\endcsname
\fi
#2}}
\providecommand{\BIBdecl}{\relax}
\BIBdecl

\bibitem{DBLP:conf/cvpr/KarpathyTSLSF14}
A.~Karpathy, G.~Toderici, S.~Shetty, T.~Leung, R.~Sukthankar, and F.~Li,
  ``Large-scale video classification with convolutional neural networks,'' in
  \emph{2014 {IEEE} Conference on Computer Vision and Pattern Recognition}, pp.
  1725--1732.

\bibitem{DBLP:conf/nips/SimonyanZ14}
K.~Simonyan and A.~Zisserman, ``Two-stream convolutional networks for action
  recognition in videos,'' in \emph{Advances in Neural Information Processing
  Systems 2014}, pp. 568--576.

\bibitem{DBLP:conf/iccv/TranBFTP15}
D.~Tran, L.~D. Bourdev, R.~Fergus, L.~Torresani, and M.~Paluri, ``Learning
  spatiotemporal features with 3d convolutional networks,'' in \emph{2015
  {IEEE} International Conference on Computer Vision}, pp. 4489--4497.

\bibitem{DBLP:journals/spl/LiuLLYY18}
Y.~Liu, Z.~Lu, J.~Li, T.~Yang, and C.~Yao, ``Global temporal representation
  based cnns for infrared action recognition,'' \emph{{IEEE} Signal Process.
  Lett.}, vol.~25, no.~6, pp. 848--852, 2018.

\bibitem{DBLP:conf/iccv/KuehneJGPS11}
H.~Kuehne, H.~Jhuang, E.~Garrote, T.~A. Poggio, and T.~Serre, ``{HMDB:} {A}
  large video database for human motion recognition,'' in \emph{2011 {IEEE}
  International Conference on Computer Vision}, pp. 2556--2563.

\bibitem{DBLP:journals/corr/abs-1212-0402}
K.~Soomro, A.~R. Zamir, and M.~Shah, ``{UCF101:} {A} dataset of 101 human
  actions classes from videos in the wild,'' \emph{CoRR}, vol. abs/1212.0402,
  2012.

\bibitem{DBLP:conf/cvpr/HeilbronEGN15}
F.~C. Heilbron, V.~Escorcia, B.~Ghanem, and J.~C. Niebles, ``Activitynet: {A}
  large-scale video benchmark for human activity understanding,'' in \emph{2015
  {IEEE} Conference on Computer Vision and Pattern Recognition}, pp. 961--970.

\bibitem{DBLP:conf/iccv/GoyalKMMWKHFYMH17}
R.~Goyal, S.~E. Kahou, V.~Michalski, J.~Materzynska, S.~Westphal, H.~Kim,
  V.~Haenel, I.~Fr{\"{u}}nd, P.~Yianilos, M.~Mueller{-}Freitag, F.~Hoppe,
  C.~Thurau, I.~Bax, and R.~Memisevic, ``The "something something" video
  database for learning and evaluating visual common sense,'' in \emph{2017
  {IEEE} International Conference on Computer Vision}, pp. 5843--5851.

\bibitem{DBLP:conf/cvpr/GuSRVPLVTRSSM18}
C.~Gu, C.~Sun, D.~A. Ross, C.~Vondrick, C.~Pantofaru, Y.~Li,
  S.~Vijayanarasimhan, G.~Toderici, S.~Ricco, R.~Sukthankar, C.~Schmid, and
  J.~Malik, ``{AVA:} {A} video dataset of spatio-temporally localized atomic
  visual actions,'' in \emph{2018 {IEEE} Conference on Computer Vision and
  Pattern Recognition}, pp. 6047--6056.

\bibitem{8651343}
M.~{Monfort}, A.~{Andonian}, B.~{Zhou}, K.~{Ramakrishnan}, S.~A. {Bargal},
  Y.~{Yan}, L.~{Brown}, Q.~{Fan}, D.~{Gutfreund}, C.~{Vondrick}, and
  A.~{Oliva}, ``Moments in time dataset: one million videos for event
  understanding,'' \emph{IEEE Transactions on Pattern Analysis and Machine
  Intelligence}, pp. 1--1, 2019.

\bibitem{DBLP:journals/pr/MaBZSS17}
S.~Ma, S.~A. Bargal, J.~Zhang, L.~Sigal, and S.~Sclaroff, ``Do less and achieve
  more: Training cnns for action recognition utilizing action images from the
  web,'' \emph{Pattern Recognition}, vol.~68, pp. 334--345, 2017.

\bibitem{DBLP:conf/iccv/YaoJKLGF11}
B.~Yao, X.~Jiang, A.~Khosla, A.~L. Lin, L.~J. Guibas, and F.~Li, ``Human action
  recognition by learning bases of action attributes and parts,'' in \emph{2011
  {IEEE} International Conference on Computer Vision}, pp. 1331--1338.

\bibitem{DBLP:journals/pami/GuptaKD09}
A.~Gupta, A.~Kembhavi, and L.~S. Davis, ``Observing human-object interactions:
  Using spatial and functional compatibility for recognition,'' \emph{{IEEE}
  Trans. Pattern Anal. Mach. Intell.}, vol.~31, no.~10, pp. 1775--1789, 2009.

\bibitem{DBLP:conf/cvpr/YaoF10}
B.~Yao and F.~Li, ``Grouplet: {A} structured image representation for
  recognizing human and object interactions,'' in \emph{2010 {IEEE} Conference
  on Computer Vision and Pattern Recognition}, pp. 9--16.

\bibitem{DBLP:conf/bmvc/DelaitreLS10}
V.~Delaitre, I.~Laptev, and J.~Sivic, ``Recognizing human actions in still
  images: a study of bag-of-features and part-based representations,'' in
  \emph{2010 British Machine Vision Conference}, pp. 1--11.

\bibitem{DBLP:conf/icpr/IkizlerCPD08}
N.~Ikizler, R.~G. Cinbis, S.~Pehlivan, and P.~Duygulu, ``Recognizing actions
  from still images,'' in \emph{2008 International Conference on Pattern
  Recognition}, pp. 1--4.

\bibitem{DBLP:conf/mm/LiWZK17}
J.~Li, Y.~Wong, Q.~Zhao, and M.~S. Kankanhalli, ``Attention transfer from web
  images for video recognition,'' in \emph{Proceedings of the 2017 {ACM} on
  Multimedia Conference}, pp. 1--9.

\bibitem{Everingham2010}
M.~Everingham, L.~Van~Gool, C.~K.~I. Williams, J.~Winn, and A.~Zisserman, ``The
  pascal visual object classes (voc) challenge,'' \emph{International Journal
  of Computer Vision}, vol.~88, no.~2, pp. 303--338, 2010.

\bibitem{DBLP:conf/ijcai/YuWSD18}
F.~Yu, X.~Wu, Y.~Sun, and L.~Duan, ``Exploiting images for video recognition
  with hierarchical generative adversarial networks,'' in \emph{2018
  International Joint Conference on Artificial Intelligence}, pp. 1107--1113.

\bibitem{DBLP:journals/tcyb/ZhangHTHJ17}
J.~Zhang, Y.~Han, J.~Tang, Q.~Hu, and J.~Jiang, ``Semi-supervised
  image-to-video adaptation for video action recognition,'' \emph{{IEEE} Trans.
  Cybernetics}, vol.~47, no.~4, pp. 960--973, 2017.

\bibitem{8578655}
Y.~{Wang}, L.~{Zhou}, and Y.~{Qiao}, ``Temporal hallucinating for action
  recognition with few still images,'' in \emph{2018 IEEE Conference on
  Computer Vision and Pattern Recognition}, pp. 5314--5322.

\bibitem{DBLP:conf/eccv/SaenkoKFD10}
K.~Saenko, B.~Kulis, M.~Fritz, and T.~Darrell, ``Adapting visual category
  models to new domains,'' in \emph{2010 European Conference on Computer
  Vision}, pp. 213--226.

\bibitem{8060555}
S.~{Ramasinghe}, J.~{Rajasegaran}, V.~{Jayasundara}, K.~{Ranasinghe},
  R.~{Rodrigo}, and A.~A. {Pasqual}, ``Combined static and motion features for
  deep-networks based activity recognition in videos,'' \emph{IEEE Transactions
  on Circuits and Systems for Video Technology}, pp. 1--1, 2017.

\bibitem{7432005}
L.~{Niu}, X.~{Xu}, L.~{Chen}, L.~{Duan}, and D.~{Xu}, ``Action and event
  recognition in videos by learning from heterogeneous web sources,''
  \emph{IEEE Transactions on Neural Networks and Learning Systems}, vol.~28,
  no.~6, pp. 1290--1304, June 2017.

\bibitem{7364249}
F.~{Liu}, X.~{Xu}, S.~{Qiu}, C.~{Qing}, and D.~{Tao}, ``Simple to complex
  transfer learning for action recognition,'' \emph{IEEE Transactions on Image
  Processing}, vol.~25, no.~2, pp. 949--960, Feb 2016.

\bibitem{DBLP:journals/complexity/0084LLYD18}
Y.~Liu, Z.~Lu, J.~Li, C.~Yao, and Y.~Deng, ``Transferable feature
  representation for visible-to-infrared cross-dataset human action
  recognition,'' \emph{Complexity}, vol. 2018, pp.
  5\,345\,241:1--5\,345\,241:20, 2018.

\bibitem{8453034}
Y.~{Liu}, Z.~{Lu}, J.~{Li}, and T.~{Yang}, ``Hierarchically learned
  view-invariant representations for cross-view action recognition,''
  \emph{IEEE Transactions on Circuits and Systems for Video Technology},
  vol.~29, no.~8, pp. 2416--2430, 2019.

\bibitem{DBLP:journals/tcsv/LuoPC09}
J.~Luo, C.~Papin, and K.~Costello, ``Towards extracting semantically meaningful
  key frames from personal video clips: From humans to computers,''
  \emph{{IEEE} Trans. Circuits Syst. Video Techn.}, vol.~19, no.~2, pp.
  289--301, 2009.

\bibitem{8578867}
D.~{Huang}, V.~{Ramanathan}, D.~{Mahajan}, L.~{Torresani}, M.~{Paluri},
  L.~{Fei-Fei}, and J.~C. {Niebles}, ``What makes a video a video: Analyzing
  temporal information in video understanding models and datasets,'' in
  \emph{2018 IEEE Conference on Computer Vision and Pattern Recognition}, pp.
  7366--7375.

\bibitem{DBLP:conf/cvpr/KodirovXG17}
E.~Kodirov, T.~Xiang, and S.~Gong, ``Semantic autoencoder for zero-shot
  learning,'' in \emph{2017 {IEEE} Conference on Computer Vision and Pattern
  Recognition}, pp. 4447--4456.

\bibitem{DBLP:conf/cvpr/LampertNH09}
C.~H. Lampert, H.~Nickisch, and S.~Harmeling, ``Learning to detect unseen
  object classes by between-class attribute transfer,'' in \emph{2009 {IEEE}
  Conference on Computer Vision and Pattern Recognition}, pp. 951--958.

\bibitem{DBLP:conf/nips/MikolovSCCD13}
T.~Mikolov, I.~Sutskever, K.~Chen, G.~S. Corrado, and J.~Dean, ``Distributed
  representations of words and phrases and their compositionality,'' in
  \emph{Advances in Neural Information Processing Systems 2013}, pp.
  3111--3119.

\bibitem{DBLP:journals/csur/AggarwalR11}
J.~K. Aggarwal and M.~S. Ryoo, ``Human activity analysis: {A} review,''
  \emph{{ACM} Comput. Surv.}, vol.~43, no.~3, pp. 16:1--16:43, 2011.

\bibitem{DBLP:conf/cvpr/LaptevMSR08}
I.~Laptev, M.~Marszalek, C.~Schmid, and B.~Rozenfeld, ``Learning realistic
  human actions from movies,'' in \emph{2008 {IEEE} Conference on Computer
  Vision and Pattern Recognition}, pp. 1--8.

\bibitem{DBLP:journals/ijcv/WangKSL13}
H.~Wang, A.~Kl{\"{a}}ser, C.~Schmid, and C.~Liu, ``Dense trajectories and
  motion boundary descriptors for action recognition,'' \emph{International
  Journal of Computer Vision}, vol. 103, no.~1, pp. 60--79, 2013.

\bibitem{DBLP:conf/iccv/WangS13a}
H.~Wang and C.~Schmid, ``Action recognition with improved trajectories,'' in
  \emph{2013 {IEEE} International Conference on Computer Vision}, pp.
  3551--3558.

\bibitem{DBLP:conf/eccv/PerronninSM10}
F.~Perronnin, J.~S{\'{a}}nchez, and T.~Mensink, ``Improving the fisher kernel
  for large-scale image classification,'' in \emph{2010 European Conference on
  Computer Vision}, pp. 143--156.

\bibitem{DBLP:conf/cvpr/JegouDSP10}
H.~Jegou, M.~Douze, C.~Schmid, and P.~P{\'{e}}rez, ``Aggregating local
  descriptors into a compact image representation,'' in \emph{2010 {IEEE}
  Conference on Computer Vision and PatternRecognition}, pp. 3304--3311.

\bibitem{DBLP:conf/cvpr/WangYYLHG10}
J.~Wang, J.~Yang, K.~Yu, F.~Lv, T.~S. Huang, and Y.~Gong,
  ``Locality-constrained linear coding for image classification,'' in
  \emph{2010 {IEEE} Conference on Computer Vision and Pattern Recognition}, pp.
  3360--3367.

\bibitem{DBLP:journals/pami/JiXYY13}
S.~Ji, W.~Xu, M.~Yang, and K.~Yu, ``3d convolutional neural networks for human
  action recognition,'' \emph{{IEEE} Trans. Pattern Anal. Mach. Intell.},
  vol.~35, no.~1, pp. 221--231, 2013.

\bibitem{DBLP:conf/cvpr/Wang0T15}
L.~Wang, Y.~Qiao, and X.~Tang, ``Action recognition with trajectory-pooled
  deep-convolutional descriptors,'' in \emph{2015 {IEEE} Conference on Computer
  Vision and Pattern Recognition}, pp. 4305--4314.

\bibitem{DBLP:conf/eccv/GanSDG16}
C.~Gan, C.~Sun, L.~Duan, and B.~Gong, ``Webly-supervised video recognition by
  mutually voting for relevant web images and web video frames,'' in \emph{2016
  European Conference on Computer Vision}, pp. 849--866.

\bibitem{DBLP:journals/tip/YuWCD19}
F.~Yu, X.~Wu, J.~Chen, and L.~Duan, ``Exploiting images for video recognition:
  Heterogeneous feature augmentation via symmetric adversarial learning,''
  \emph{{IEEE} Trans. Image Processing}, vol.~28, no.~11, pp. 5308--5321, 2019.

\bibitem{DBLP:conf/iccv/LongWDSY13}
M.~Long, J.~Wang, G.~Ding, J.~Sun, and P.~S. Yu, ``Transfer feature learning
  with joint distribution adaptation,'' in \emph{2013 {IEEE} International
  Conference on Computer Vision}, pp. 2200--2207.

\bibitem{DBLP:conf/cvpr/ZhangLO17}
J.~Zhang, W.~Li, and P.~Ogunbona, ``Joint geometrical and statistical alignment
  for visual domain adaptation,'' in \emph{2017 {IEEE} Conference on Computer
  Vision and Pattern Recognition}, pp. 5150--5158.

\bibitem{DBLP:conf/icml/LongC0J15}
M.~Long, Y.~Cao, J.~Wang, and M.~I. Jordan, ``Learning transferable features
  with deep adaptation networks,'' in \emph{Proceedings of the 2015
  International Conference on Machine Learning}, pp. 97--105.

\bibitem{8502831}
C.~{Zhang}, H.~{Fu}, Q.~{Hu}, X.~{Cao}, Y.~{Xie}, D.~{Tao}, and D.~{Xu},
  ``Generalized latent multi-view subspace clustering,'' \emph{IEEE
  Transactions on Pattern Analysis and Machine Intelligence}, pp. 1--1, 2018.

\bibitem{DBLP:conf/cvpr/JiangL17}
Q.~Jiang and W.~Li, ``Deep cross-modal hashing,'' in \emph{2017 {IEEE}
  Conference on Computer Vision and Pattern Recognition}, pp. 3270--3278.

\bibitem{DBLP:conf/cvpr/HuangP18}
X.~Huang and Y.~Peng, ``Deep cross-media knowledge transfer,'' in \emph{2018
  {IEEE} Conference on Computer Vision and Pattern Recognition}, pp.
  8837--8846.

\bibitem{DBLP:journals/tist/ChangL11}
C.~Chang and C.~Lin, ``{LIBSVM:} {A} library for support vector machines,''
  \emph{{ACM} {TIST}}, vol.~2, no.~3, pp. 27:1--27:27, 2011.

\bibitem{DBLP:conf/bmvc/ChatfieldSVZ14}
K.~Chatfield, K.~Simonyan, A.~Vedaldi, and A.~Zisserman, ``Return of the devil
  in the details: Delving deep into convolutional nets,'' in \emph{2014 British
  Machine Vision Conference}.

\bibitem{DBLP:conf/nips/KrizhevskySH12}
A.~Krizhevsky, I.~Sutskever, and G.~E. Hinton, ``Imagenet classification with
  deep convolutional neural networks,'' in \emph{Advances in Neural Information
  Processing Systems 2012}, pp. 1106--1114.

\bibitem{DBLP:conf/cvpr/VenkateswaraECP17}
H.~Venkateswara, J.~Eusebio, S.~Chakraborty, and S.~Panchanathan, ``Deep
  hashing network for unsupervised domain adaptation,'' in \emph{2017 {IEEE}
  Conference on Computer Vision and Pattern Recognition}, pp. 5385--5394.

\bibitem{DBLP:conf/iccv/CaoLWY17}
Z.~Cao, M.~Long, J.~Wang, and P.~S. Yu, ``Hashnet: Deep learning to hash by
  continuation,'' in \emph{2017 {IEEE} International Conference on Computer
  Vision}, pp. 5609--5618.

\bibitem{DBLP:conf/nips/RanzatoBL07}
M.~Ranzato, Y.~Boureau, and Y.~LeCun, ``Sparse feature learning for deep belief
  networks,'' in \emph{Advances in Neural Information Processing Systems 2007},
  pp. 1185--1192.

\bibitem{DBLP:journals/cacm/BartelsS72}
R.~H. Bartels and G.~W. Stewart, ``Solution of the matrix equation ax+xb=c
  {[F4]} (algorithm 432),'' \emph{Commun. {ACM}}, vol.~15, no.~9, pp. 820--826,
  1972.

\bibitem{DBLP:journals/ijcv/RussakovskyDSKS15}
O.~Russakovsky, J.~Deng, H.~Su, J.~Krause, S.~Satheesh, S.~Ma, Z.~Huang,
  A.~Karpathy, A.~Khosla, M.~S. Bernstein, A.~C. Berg, and F.~Li, ``Imagenet
  large scale visual recognition challenge,'' \emph{International Journal of
  Computer Vision}, vol. 115, no.~3, pp. 211--252, 2015.

\bibitem{DBLP:journals/ijcv/ZhuS14}
F.~Zhu and L.~Shao, ``Weakly-supervised cross-domain dictionary learning for
  visual recognition,'' \emph{International Journal of Computer Vision}, vol.
  109, no. 1-2, pp. 42--59, 2014.

\bibitem{DBLP:conf/emnlp/MilajevsKSP14}
D.~Milajevs, D.~Kartsaklis, M.~Sadrzadeh, and M.~Purver, ``Evaluating neural
  word representations in tensor-based compositional settings,'' in
  \emph{Proceedings of the 2014 Conference on Empirical Methods in Natural
  Language Processing}, pp. 708--719.

\bibitem{DBLP:journals/tnn/PanTKY11}
S.~J. Pan, I.~W. Tsang, J.~T. Kwok, and Q.~Yang, ``Domain adaptation via
  transfer component analysis,'' \emph{{IEEE} Trans. Neural Networks}, vol.~22,
  no.~2, pp. 199--210, 2011.

\bibitem{DBLP:conf/cvpr/GongSSG12}
B.~Gong, Y.~Shi, F.~Sha, and K.~Grauman, ``Geodesic flow kernel for
  unsupervised domain adaptation,'' in \emph{2012 {IEEE} Conference on Computer
  Vision and Pattern Recognition}, pp. 2066--2073.

\bibitem{DBLP:conf/cvpr/LongWDSY14}
M.~Long, J.~Wang, G.~Ding, J.~Sun, and P.~S. Yu, ``Transfer joint matching for
  unsupervised domain adaptation,'' in \emph{2014 {IEEE} Conference on Computer
  Vision and Pattern Recognition}, pp. 1410--1417.

\bibitem{DBLP:conf/aaai/SunFS16}
B.~Sun, J.~Feng, and K.~Saenko, ``Return of frustratingly easy domain
  adaptation,'' in \emph{2016 Proceedings of the Thirtieth {AAAI} Conference on
  Artificial Intelligence}, pp. 2058--2065.

\bibitem{DBLP:conf/mm/WangFCYHY18}
J.~Wang, W.~Feng, Y.~Chen, H.~Yu, M.~Huang, and P.~S. Yu, ``Visual domain
  adaptation with manifold embedded distribution alignment,'' in \emph{2018
  {ACM} Multimedia Conference on Multimedia Conference}, pp. 402--410.

\bibitem{DBLP:conf/cvpr/BilenFGVG16}
H.~Bilen, B.~Fernando, E.~Gavves, A.~Vedaldi, and S.~Gould, ``Dynamic image
  networks for action recognition,'' in \emph{2016 {IEEE} Conference on
  Computer Vision and Pattern Recognition}, pp. 3034--3042.

\bibitem{DBLP:journals/pami/DonahueHRVGSD17}
J.~Donahue, L.~A. Hendricks, M.~Rohrbach, S.~Venugopalan, S.~Guadarrama,
  K.~Saenko, and T.~Darrell, ``Long-term recurrent convolutional networks for
  visual recognition and description,'' \emph{{IEEE} Trans. Pattern Anal. Mach.
  Intell.}, vol.~39, no.~4, pp. 677--691, 2017.

\bibitem{DBLP:conf/icml/SrivastavaMS15}
N.~Srivastava, E.~Mansimov, and R.~Salakhutdinov, ``Unsupervised learning of
  video representations using lstms,'' in \emph{2015 International Conference
  on Machine Learning}, pp. 843--852.

\bibitem{DBLP:conf/iccv/SunJYS15}
L.~Sun, K.~Jia, D.~Yeung, and B.~E. Shi, ``Human action recognition using
  factorized spatio-temporal convolutional networks,'' in \emph{2015 {IEEE}
  International Conference on Computer Vision}, pp. 4597--4605.

\bibitem{DBLP:journals/pami/VarolLS18}
G.~Varol, I.~Laptev, and C.~Schmid, ``Long-term temporal convolutions for
  action recognition,'' \emph{{IEEE} Trans. Pattern Anal. Mach. Intell.},
  vol.~40, no.~6, pp. 1510--1517, 2018.

\bibitem{DBLP:conf/cvpr/ZhuHSCQ16}
W.~Zhu, J.~Hu, G.~Sun, X.~Cao, and Y.~Qiao, ``A key volume mining deep
  framework for action recognition,'' in \emph{2016 {IEEE} Conference on
  Computer Vision and Pattern Recognition}, pp. 1991--1999.

\bibitem{DBLP:conf/eccv/WangXW0LTG16}
L.~Wang, Y.~Xiong, Z.~Wang, Y.~Qiao, D.~Lin, X.~Tang, and L.~V. Gool,
  ``Temporal segment networks: Towards good practices for deep action
  recognition,'' in \emph{2016 European Conference on Computer Vision}, pp.
  20--36.

\bibitem{DBLP:conf/ifip12/ZangWLZHZ18}
J.~Zang, L.~Wang, Z.~Liu, Q.~Zhang, G.~Hua, and N.~Zheng, ``Attention-based
  temporal weighted convolutional neural network for action recognition,'' in
  \emph{2018 Artificial Intelligence Applications and Innovations}, pp.
  97--108.

\bibitem{DBLP:conf/cvpr/CarreiraZ17}
J.~Carreira and A.~Zisserman, ``Quo vadis, action recognition? {A} new model
  and the kinetics dataset,'' in \emph{2017 {IEEE} Conference on Computer
  Vision and Pattern Recognition}, pp. 4724--4733.

\end{thebibliography}

\end{document}